%% file: main.tex
\documentclass[10pt,twocolumn,letterpaper]{article}

\usepackage{cvpr}

\input{preamble}

\definecolor{cvprblue}{rgb}{0.21,0.49,0.74}
\usepackage[pagebackref,breaklinks,colorlinks,citecolor=cvprblue]{hyperref}
\newcommand{\parag}[1]{\noindent\emph{#1}}
\newcommand{\cropImg}{I_{\text{crop}}}
\newcommand{\hoiImg}{I_{\text{hoi}}}
\newcommand{\mesh}{\mathcal{M}}
\newcommand{\objImg}{I_{\text{obj}}}
\newcommand{\Pm}{P_{\text{hoi}}}

\newcommand{\Hhand}{\mathcal{M}^{H}_h}
\newcommand{\Ihand}{\mathcal{M}^{I}_h}
\newcommand{\Iobj}{\mathcal{M}^{I}_o}

\newcommand{\objM}{\widetilde O} 
\newcommand{\handM}{\widetilde H} 
\newcommand{\hoiM}{\widetilde {HOI}} 
\newcommand{\UtoI}{\mathcal{T}_{U \rightarrow I}} 
\newcommand{\HtoI}{\mathcal{T}_{H \rightarrow I}} 
\newcommand{\velocity}{\mathbf{v}_\theta(x_t, t,\mathbf{c})}  
\newcommand{\velobj}{\mathbf{v}_\theta(x_t, t,\mathbf{c}_{obj})}  
\newcommand{\perturbedVelobj}{{\mathbf{v}}^{*}_\theta(x_t, t, \mathbf{c})}  
\newcommand{\latent}{\mathbf{x}}

\title{Follow My Hold: Hand-Object Interaction Reconstruction through Geometric Guidance}

\author{Ayce Idil Aytekin$^1$\thanks{Corresponding author: \texttt{aaytekin@mpi-inf.mpg.de}}\quad
Helge Rhodin$^2$\quad
Rishabh Dabral$^1$\quad
Christian Theobalt$^1$\\
$^1$ Max Planck Institute for Informatics and Saarland University \\
$^2$ Bielefeld University\\
{\small \url{https://aidilayce.github.io/FollowMyHold-page/}}
}

\begin{document}

\maketitle
\input{fig/teaser}
\input{sec/0_abstract}    
\input{sec/1_intro}
\input{sec/2_related_work}

\input{sec/3_preliminary}
\input{sec/4_method}
\input{sec/5_experiments}

\input{sec/6_conclusion}

{
    \small
    \bibliographystyle{ieeenat_fullname}
    \bibliography{references}
}
\input{sec/7_supplementary}
\end{document}

%% file: preamble.tex
\usepackage[dvipsnames]{xcolor}
\usepackage{graphicx}
\usepackage{multirow}
\usepackage{pifont}
\usepackage{algorithm}
\usepackage{algpseudocode}
\usepackage[normalem]{ulem}
\usepackage{amsmath}
\usepackage{amsthm}

\usepackage{wrapfig}
\usepackage{colortbl}
\usepackage[utf8x]{inputenc}
\usepackage{cuted}
\usepackage{bbm}



%% file: fig/teaser.tex
\begin{strip}
	\centering
     \includegraphics[width=1.0\linewidth]{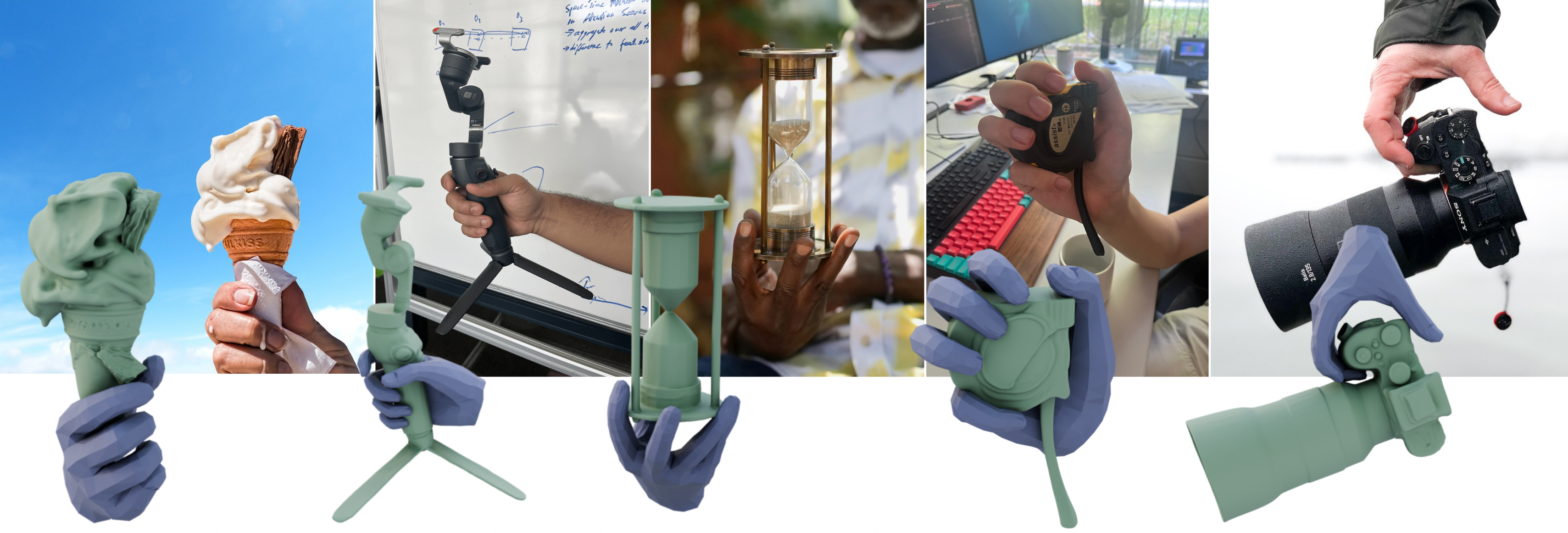}
	\captionof{figure}{ 
        \textbf{In-the-wild results.}
        From a single RGB frame from in-the-wild (top row), our method reconstructs detailed 3D hand–object interactions (bottom row).
        It does so by guiding a latent diffusion model with multi-modal cues derived from the input image and several foundation models.
    }
    \label{fig:teaser}
	\vspace{-5pt}
\end{strip}

%% file: sec/0_abstract.tex
\begin{abstract}
We propose a novel diffusion-based framework for reconstructing 3D geometry of hand-held objects from monocular RGB images by leveraging hand-object interaction as geometric guidance. 
Our method conditions a latent diffusion model on an inpainted object appearance and uses inference-time guidance to optimize the object reconstruction, while simultaneously ensuring plausible hand-object interactions.
Unlike prior methods that rely on extensive post-processing or produce low-quality reconstructions, our approach directly generates high-quality object geometry during the diffusion process by introducing guidance with an optimization-in-the-loop design.
Specifically, we guide the diffusion model by applying supervision to the velocity field while simultaneously optimizing the transformations of both the hand and the object being reconstructed.
This optimization is driven by multi-modal geometric cues, including normal and depth alignment, silhouette consistency, and 2D keypoint reprojection.
We further incorporate signed distance field supervision and enforce contact and non-intersection constraints to ensure physical plausibility of hand-object interaction. 
Our method yields accurate, robust and coherent reconstructions under occlusion while generalizing well to in-the-wild scenarios.
%
%
\end{abstract}
\vspace{-10pt}

%% file: sec/1_intro.tex
\section{Introduction}
\label{sec:intro}
%
%
Our hands are how we shape the world; by picking, pushing, holding, slicing, or sculpting. 
From watering flowers to assembling furniture, hand-object interactions are ubiquitous.
However, from a 3D reconstruction perspective, these interactions are inherently ambiguous: hands occlude objects and shapes overlap.
This makes an accurate 3D reconstruction of hand-object interaction (HOI) difficult as ambiguity is even greater when working with a single image, where depth and contact cues are limited.
Tackling this practical setting is crucial to enable robust and scalable 3D understanding in AR/VR, embodied AI, and robotics.
\par 
It is noteworthy that hand and object reconstruction have been long-standing problems~\cite{corona2020ganhand, grady2021contactopt, hasson2019learning, hasson2021towards, liu2021semi, tekin2019h+, tse2022collaborative} with several challenges arising out of the ill-posedness of the task.
Unlike clothed bodies or textured objects, hands are of rather homogeneous colors, making it challenging to reconstruct them from a single image.
Likewise, reconstructing objects in 3D, despite the impressive progress in the recent years \cite{what3d_cvpr19, umr2020, GRAF, xiang2024structured, hunyuan}, remains a challenging task since objects come in such diverse shapes that building a strong prior is difficult.
The reconstruction problem becomes significantly more complex when we combine the two tasks into one: reconstructing the 3D objects when they are undergoing hand-held interaction.
\par
Needless to say, one cannot solve the HOI reconstruction problem without reasonably modeling the occlusion caused by the hands. 
In this line, existing works have attempted several approaches, such as template-based optimization \cite{gkioxari2018detecting, grady2021contactopt, hamer2010object, hampali2020honnotate, zhang2020perceiving}, training on 3D hand-object data \cite{hasson2019learning, karunratanakul2020grasping, ye2022ihoi}, data-driven normal prediction \cite{fan2024hold} or 6DoF prediction \cite{tekin2019h+, cao2021reconstructing}.
However, such models either require explicit geometric priors in the form of a template or are better suited for reconstruction with multiple views/frames.
Moreover, methods that perform a direct regression of the partial point-clouds~\cite{moge} lack the ability to complete the unobserved portions of the captured object.
\par
Recently, methods like EasyHOI \cite{liu2024easyhoi} and Gen3DSR \cite{dogaru2024generalizable} have attempted to leverage the large-scale 2D and 3D foundational models to propose pipelines that combine the implicit priors in different foundation models into one.
Despite impressive out-of-domain generalizability, they tend to be brittle, and failure of a single component during optimization often results in a catastrophic failure of the reconstruction.
Instead of directly adopting a generative model's output as the final reconstruction, we propose to embed 2D foundational model supervision directly into the 3D generative sampling process to improve robustness.
\par
With this core idea, we introduce \textbf{FollowMyHold}, which \emph{guides} a pretrained flow-based image-to-3D generator \emph{during} inference with geometric supervision. 
Concretely, we (i) extract complementary cues with 2D/3D foundation models: interaction masks~\cite{langsam}, hand detection~\cite{wilor}, inpainted object appearance~\cite{labs2025flux1kontextflowmatching}, an initial hand mesh~\cite{hamer}, an initial coarse HOI mesh from Hunyuan3D-2~\cite{hunyuan}, and a partial HOI point cloud~\cite{moge}; (ii) register all outputs into a shared image-aligned frame; and (iii) steer a rectified-flow 3D generator (Hunyuan3D-2) with an optimization-in-the-loop design for sampling.
%
Our staged optimization (hand, object, then joint refinement) applies pixel-aligned 2D losses (normal, depth, silhouette, keypoints) together with 3D interaction constraints (intersection and proximity), yielding physically plausible and consistent HOI reconstructions.
\par
We evaluate on well-established HOI image-to-3D reconstruction benchmarks like OakInk~\cite{yang2022oakink}, Arctic~\cite{fan2023arctic} and DexYCB~\cite{chao2021dexycb}.
%
%
FollowMyHold sets the state-of-the-art among generative methods, surpassing EasyHOI in object reconstruction accuracy and achieving almost two-fold higher reconstruction rate, with strong robustness and in-the-wild generalization.
Extensive evaluation validates the accuracy and robustness of our method.
The code and demo will be available for research.
%
%

%% file: sec/2_related_work.tex
\section{Related Work}
\label{sec:related_work}
\subsection{Reconstructing Hand-Object Interactions}
Recovering the 3D structure of interacting hands and objects from a single image is challenging due to occlusions and limited 3D supervision.
Many existing methods simplify the problem by assuming access to 3D object templates \cite{brahmbhatt2020contactpose, cao2021reconstructing, chen2021joint, corona2020ganhand, garcia2018first, gkioxari2018detecting, grady2021contactopt, hamer2010object, hampali2020honnotate, zhang2020perceiving}, estimating only the hand and object poses from video sequences \cite{patel2022learningimitateobjectinteractions, chen2022tracking} or multi-view inputs \cite{Qu_2023_ICCV}.
Template-free approaches learn from 3D interaction datasets \cite{chen2022alignsdf, chen2023gsdf} but often struggle due to limited object diversity in their training set. 
Recent work \cite{liu2024easyhoi} uses foundation models to reconstruct hands and objects separately but fixes object geometry early, making the pipeline brittle to initial errors.
In contrast, our method jointly optimizes object shape, pose, and hand pose in a category-agnostic manner, leveraging foundation model priors while guiding the model via geometric signals.
\subsection{Monocular 3D Object Reconstruction}
Recovering 3D shape from a single image remains one of the most fundamental problems in 3D vision. 
From early CNN-based models \cite{umr2020, GRAF} to retrieval-based systems \cite{what3d_cvpr19} and volumetric methods \cite{GIRAFFE}, a wide array of strategies have been proposed.
Recent trends shift toward generative reconstruction, with Large Reconstruction Models (LRMs) \cite{hong2023lrm, tang2024lgm, wang2023pf, wang2024crm, xiang2024structured, hunyuan} enabling high-fidelity geometry from sparse input.
However, these models are typically trained under assumptions of object-centric, unobstructed inputs.
In practice, reconstructions degrade sharply in the presence of occlusion, such as during human-object interaction. 
Our approach adapts large-scale object reconstruction models to this more complex setting by integrating additional cues derived from hand-object contact and visibility, guiding them toward more plausible completions even under partial observation.
\subsection{Hand Mesh Recovery}
Estimating hand pose and shape from images has seen major progress with the introduction of parametric models like MANO, allowing dense predictions from monocular RGB \cite{hamer, wilor, hamba}. 
Recent approaches either directly regress MANO parameters \cite{baek2019pushing, boukhayma20193d} or optimize them via image-level constraints \cite{zhang2019end, zhou2020monocular}. 
While these techniques perform well in isolation, they often fail to account for physical plausibility in the presence of interacting objects.
In our method, we employ HaMeR \cite{hamer}, a state-of-the-art hand mesh recovery method, and use this mesh as a signal to guide the object reconstruction process.

%% file: sec/3_preliminary.tex
\section{Preliminaries: Rectified Flow and Notation}
\label{sec:prelim}
We review rectified flow model, which is the backbone of our 3D generator (Hunyuan3D-2), to clarify how inference-time guidance operates.
\par
\noindent \textbf{Rectified flow {$\hat{\textbf{x}}_1$} formulation.} 
Rectified flow models are a class of generative models trained with the flow-matching objective~\cite{flowmatching}, framing generation as solving an ordinary differential equation (ODE).
Instead of predicting noise (as in DDPMs, which approximate the score function), the model learns a \emph{velocity field} that moves a noisy sample toward a clean target over time.
%
Given a noisy sample $\latent_t$ at time $t$ and conditioning $\mathbf{c}$, the model predicts $\velocity$, and the update is
\begin{equation}
\label{eqn:diffusion_update}
\latent_{t+\Delta t} \;=\; \latent_t \;+\; (\sigma_{t+\Delta t}-\sigma_t)\, \velocity,
\end{equation}
where $\sigma_t \in [0,1]$ encodes the flow time at step $t \in [0,1]$.
At inference, we \emph{steer} generation without changing the model weights by modifying $\velocity$ using gradients of a task objective $G$ (Sec.~\ref{sec:method}), e.g., geometry-consistency losses, to nudge the trajectory toward lower $G$.
To evaluate $G$ we often need the estimate of the clean sample $\hat \latent_1$ (following Hunyuan3D-2's $\latent_{1}$-target formulation), as it is the current approximation of the underlying target. 
This is recovered from the flow $\velocity$ using Eq. 4.54 in~\cite{lipman2024flow}, 
\begin{equation}
\label{eq:get_Z1}
\hat{\latent}_1 \;=\; \latent_t \;+\; (1-\sigma_t)\, \velocity.
\end{equation}
See \cref{supp:clean_sample_estimate} of the supplementary for derivation details.
\par 
\noindent \textbf{Notation.} 
We denote meshes by $\mathcal{M}$, with superscripts indicating their canonical spaces:  
\( \mathcal{M}^{\text{U}} \) for H\textbf{u}nyuan, \( \mathcal{M}^{\text{H}} \) for \textbf{H}aMeR, and \( \mathcal{M}^{\text{I}} \) for \textbf{i}mage-aligned space. 
Transformations between spaces are represented by a similarity transform $\mathcal{T}$, parameterized by scale \( s \in \mathbb{R} \), rotation \( R \in \mathrm{SO}(3) \) and the translation \( \mathbf{t} \in \mathbb{R}^3 \).
For example, a mesh in HaMeR's output space could be transformed to image-aligned space via $\mathcal{M}^{\text{I}} = \mathcal{T}_{\text{H} \rightarrow \text{I}} \cdot \mathcal{M}^{\text{H}}$.

%% file: sec/4_method.tex
\section{Method}
\label{sec:method}
\input{fig/overview}
Reconstructing a physically plausible 3D hand-object interaction from a single RGB image is a fundamentally ill-posed task.
Occlusion, entangled geometry, and limited depth cues make it difficult to recover accurate hand and object shapes, and consequently their spatial arrangement.
Recent foundation models offer strong priors for hands (e.g., HaMeR), 3D geometry (e.g., Hunyuan3D-2), and pixel-aligned partial geometry with depth cues (e.g., MoGe-2).
Each model presents a set of complementary strengths and weaknesses, and our goal is to propose a method that effectively navigates these conflicting properties.
However, the methods are incompatible, e.g., they operate in their own canonical coordinate space, with mismatching assumptions about scale, orientation, and alignment.
\par
Simply connecting the outputs of these models, as attempted by recent methods like EasyHOI~\cite{liu2024easyhoi} and Gen3DSR~\cite{dogaru2024generalizable}, does not work well as the misalignment of canonical spaces prevents direct comparison or joint optimization of model outputs.
Our method directly tackles this coordination challenge by explicitly aligning model outputs into a shared, \textit{image-aligned} reference frame.
This alignment unlocks a crucial capability: we can now compare the rendered normals, disparity, and silhouette maps from our 3D predictions directly against 2D supervision maps rendered from MoGe-2~\cite{moge}, a state-of-the-art method for point-cloud prediction.
This enables pixel-accurate, differentiable supervision throughout the optimization process.
\par
At a high level, our method (1) uses foundation models to segment and inpaint the image and produce initial 3D predictions, including a hand mesh, a partial point cloud of HOI, and a coarse HOI mesh (Sec.~\ref{subsubsec:preprocessing2d} and Sec. \ref{subsubsec:preprocessing3d}), (2) aligns all outputs to a shared coordinate frame via a two-stage transformation chain (Sec.~\ref{subsubsec:icp}), and (3) renders from this unified frame and applies gradient-based guidance during diffusion using both 2D supervision (e.g., normals, depth, silhouette) and 3D interaction constraints (e.g., intersection and proximity losses) (Sec.~\ref{sec:guidance}).
See Fig. \ref{fig:overview} for an overview of our method.

\subsection{Initialization with Foundational Models}
\label{subsec:init}
\subsubsection{2D Signal Extraction}
\label{subsubsec:preprocessing2d}
Given as input a single RGB image $I_{\text{full}}\in\mathbb{R}^{\hat H \times \hat W \times 3}$, we first localize the interaction region using foundation models. 
Specifically, LangSAM \cite{langsam} is used to extract the hand ($M_h$) and object ($M_o$) masks, while WiLoR's hand detector \cite{wilor} detects hand bounding boxes.
Together they define a crop yielding $\cropImg\!\in\!\mathbb{R}^{H \times W \times 3}$. 
We then use a diffusion-based inpainter (FLUX.1 Kontext[dev]~\cite{labs2025flux1kontextflowmatching} guided by a Gemini~\cite{gemini} text prompt) to remove the hand and complete the occluded object appearance, producing the inpainted object image $\objImg$.
We additionally mask $\cropImg$ with the union of $M_h$ and $M_o$ to obtain $I_{\text{hoi}}$, which serves as input to MoGe-2 and Hunyuan3D-2.
\vspace{-5pt}
\subsubsection{3D Geometry Initialization}
\label{subsubsec:preprocessing3d}
We precompute 3D cues to guide diffusion-based HOI reconstruction.
To begin with, we employ a state-of-the-art hand-reconstruction model, HaMeR \cite{hamer}, that provides an initial hand mesh $\mesh^H_h$ and its 2D keypoints $\hat K_{2D}$ from $\cropImg$.
Then, MoGe-2 \cite{moge} estimates the partial point cloud $P_{\text{hoi}}\in\mathbb{R}^{N_m\times3}$ and camera $\phi$ from $\hoiImg$.
The partial point cloud $P_{\text{hoi}}$ output by MoGe-2 only consists of the points corresponding to the unoccluded portions of the input object image $I_{\text{hoi}}$, as shown in~\cref{fig:overview}.
We then render the normal map, disparity map, and silhouette from \( \Pm \) using $\phi$ to serve as the supervision maps for the geometric guidance objective during the sampling process.
\par
At the core of our framework lies Hunyuan3D-2~\cite{hunyuan}, a large-scale latent diffusion model for 3D shape generation.
The diffusion transformer $\mathcal{E}_{\text{hoi}}$ produces the HOI latent $\latent_{\text{hoi}}=\mathcal{E}_{\text{hoi}}(\latent_t; t, \mathbf{c}_{\text{hoi}})$, where $\mathbf{c}_{\text{hoi}}$ are DINOv2~\cite{dinov2} features of $\cropImg$. 
The generated latent $\latent_{\text{hoi}}$ is decoded into a signed distance field (SDF) by Hunyuan3D-2's decoder $\mathcal{D}$ over a query grid $\mathbf{q}\in [-1,1]^3$, and meshed via FlexiCubes~\cite{flexicubes}: $
\mathrm{SDF}_{\text{hoi}} = \mathcal{D}(\mathbf{x}_{\text{hoi}}, q)$ and $
\mesh^{U}_{\text{hoi}} = \mathrm{FlexiCubes}(\mathrm{SDF}_{\text{hoi}}).
$
Although trained on single objects, Hunyuan3D-2 preserves coarse hand–object arrangements (see supplemental material, \cref{fig:coarse_hunyuan}), which we leverage as cues for transforming and aligning meshes across canonical spaces.
\subsubsection{Transforming Across Canonical Spaces with ICP}
\label{subsubsec:icp}
While critical to our pipeline, the outputs from HaMeR, Hunyuan3D-2, and MoGe-2 lie in different coordinate spaces.
To enable joint reasoning and rendering, we estimate rigid transformations between these coordinate spaces using Iterative Closest Point (ICP) alignment.
\par%
We first align the coarse HOI mesh to the partial point cloud (yielding $T_{U\to I}$), then align the hand mesh to the coarse HOI mesh (resulting in $T_{H\to U}$) to compose them and obtain $T_{H\to I}$; this two-step ICP avoids the pitfalls of directly aligning the hand to an incomplete point cloud.
These transformations bring all the geometric entities into a unified coordinate frame (image frame of MoGe-2 partial point cloud) for consistent modeling and inference-time guidance. 
Note that we initialize ICP with a similarity transformation that aligns the centroids and global scales of the source and the target.
\subsection{Inference-Time Guidance and Staged Optimization}
We combine inference-time guidance with staged optimization of transformation parameters.  
We first explain our guidance strategy in rectified flow and then the role of each optimization phase.
\subsubsection{Inference-time Guidance}
\label{sec:guidance}
Our base image-to-3D generative model, Hunyuan3D-2, uses a rectified flow model (Sec. \ref{sec:prelim}) designed for efficient image-to-shape generation.
We modify this model by integrating gradient-based supervision directly into the flow trajectory at inference time, enabling end-to-end refinement of object geometry and pose under loss-specific constraints.
We structure our reconstruction process into three phases, each addressing specific challenges in HOI recovery.
\par
Before introducing our geometric guidance, we let the denoising process reach time step $t \geq \tau_1$, at which the denoised latents reach a coarse but sufficient quality.
For early diffusion steps $t < \tau_{1}$, we apply standard rectified flow updates in Eq. \ref{eqn:diffusion_update} without guidance. 
\input{tab/algorithm}
Algorithm~\ref{algo:opt} describes our full inference-time multi-step optimization used in Phase 3, where both the object and hand transformations $T_o$ and $T_h$ are jointly optimized and the velocity field $\velobj$ is guided using both 2D and 3D objectives.
We denote the flow field $\velocity$ as $\mathbf{v}_t$ for simplicity.
$\mathcal{R}_s$ represents supervision maps rendered from $P_{\text{hoi}}$: normal, disparity, and silhouette maps. 
Unlike traditional inference-time guidance, we follow an optimization-in-the-loop strategy, jointly optimizing the transformations ($T_h$ and $T_o$) along with the flow field estimated by the model.
In particular, for each diffusion time step $t \in [ \tau_2, 1]$, we perform $K$ inner gradient descent iterations to jointly optimize the transformations and steer $\velocity$ using 2D and 3D geometric objectives $G_{2D}$ and $G_{3D}$.
\par
Across phases, 2D losses compare rendered maps against \(\mathcal{R}_s\): normal alignment \(L_{\text{norm},(\cdot)}\), L1 disparity \(L_{\text{disp},(\cdot)}\), and binary cross-entropy silhouette \(L_{\text{sil},(\cdot)}\), with \((\cdot)\in\{\text{h},\text{o},\text{hoi}\}\) for hand, object, and their union.
Note that we treat the hand geometry as fixed during object reconstruction, as HaMeR provides reliable hand estimates even under occlusion. 
In contrast, the object is more underconstrained due to partial visibility and the wide range of plausible shapes.  
We use the hand as a stable geometric anchor to constrain object pose optimization.
Overall, this staged strategy enables stable and physically plausible 3D interaction reconstruction from a single RGB image.
The phases are explained in the following.
\subsubsection{Phase 1: Hand-Only Optimization}
Phase 1 only focuses on optimizing the hand to establish a spatially meaningful anchor before introducing guidance on the object.
At diffusion step \( t = \tau_{1} \), we optimize the hand transformation \( T_h \), initialized with unit scale, identity rotation, and zero translation, while keeping the object transformation parameters \( T_o \) and the velocity field \(\velobj\) fixed.
We first transform the MANO hand mesh $\Hhand$ into the image-aligned space as $\Ihand = \HtoI \Hhand$ and pose it with $\widetilde H = T_h \cdot \Ihand$.
%
Rendering \(\widetilde{H}\) yields \(L_{\text{norm,h}},L_{\text{disp,h}},L_{\text{sil,h}}\). 
We also add the $\ell_2$ loss between the projected 3D hand keypoints and their corresponding 2D keypoints detected by HaMeR and regularize excessive translation using $L_{\text{reg,h}} = \bigl\lVert \mathbf{t}_h \bigr\rVert^2$.
%
%
The Phase 1 objective \(L_{\text{phase1}}\) sums these terms (weights in supplemental material~\cref{supp:implementation}). 
Once complete, we obtain a well-posed hand mesh and proceed with the denoising process.
%
%
\subsubsection{Phase 2: Object-Only Optimization}
This phase tackles a robustness issue: if the object reconstructed at \( t = \tau_1 \) is heavily misaligned or poorly scaled, it can occlude or be occluded by the hand in the rendered views, weakening 2D loss supervision.
To mitigate this, we coarsely align the object before enabling joint hand-object reasoning in Phase 3.
At \( t = \tau_2 \), we optimize \( T_o \) and apply gradient-based guidance to \( \velobj \), while keeping the hand transformation \( T_h \) fixed.
The operations in this phase are similar to Algorithm \ref{algo:opt} without the hand mesh. 
\noindent \( T_o \) is initialized as identity, and following Sec.~\ref{sec:guidance}, we extract the object's mesh, which is defined in the canonical space of Hunyuan3D-2.
We then transform it to image-aligned space via $\Iobj = \UtoI \Iobj$ and pose it by $\objM = T_o \cdot \Iobj$.
From \( \objM \), we render geometry maps and compute the supervision losses \( L_{\text{norm,o}}, L_{\text{disp,o}}, L_{\text{sil,o}} \).
A regularization loss \( L_{\text{reg,o}} \) is computed to penalize excessive translation, scale drift, and mesh noise.
The Phase 2 loss \(L_{\text{phase2}}\) (sum of these terms; weights in supplemental material in~\cref{supp:implementation}) serves as $G_{2D}$ to steer \(\velobj\) while updating \(T_o\) for \(k_2\) steps. 
Once complete, we obtain a coarsely posed object mesh and a supervision-steered velocity field $\perturbedVelobj$, and proceed by updating the object latent using Eq.~\ref{eqn:diffusion_update}.
%
%
\subsubsection{Phase 3: Joint Optimization}
In this final phase (at $t \in [\tau_2, 1]$), we jointly refine \(\velobj\), \( T_o \), and \(T_h\).
This phase introduces 3D geometric supervision alongside 2D objectives to ensure spatial plausibility and physical interaction.
\par
\noindent \textbf{2D geometric supervision:} 
With \(\handM\) and \(\objM\) posed, we define $\hoiM = \handM \cup \objM$ and compute \(L_{\text{norm,hoi}},L_{\text{disp,hoi}},L_{\text{sil,hoi}}\).
\noindent \textbf{3D geometric supervision:} 
To ensure physically plausible hand-object interactions, we introduce intersection and proximity losses.
\par
\parag{Intersection loss.} 
We add an intersection penalty with the goal of introducing a guidance objective that nudges the sampling process towards a 3D latent that does not intersect with the hand.
Specifically, we convert the posed hand and object meshes into signed distance fields (SDFs) on a shared volumetric grid and define the objective as
\begin{equation}
L_{\text{intersection}} = \tfrac{1}{K} \sum_{i=1}^{K} 
\mathbbm{1}\!\left[\text{SDF}_{h}(f_i) < 0 \wedge \text{SDF}_{o}(f_i) < 0\right],
\end{equation}
where \(\mathbbm{1} [ \cdot ]\) represents the indicator function.
This loss penalizes regions where both SDFs are negative, indicating a volumetric overlap.
\par
\parag{Proximity loss.} 
While the intersection term ensures that the resulting \(\text{SDF}_{\text{o}}\) does not penetrate the hands, it does not guarantee that the object stays in proximity to the hand.
Hence, we define the proximity loss using attraction terms with margin \(\delta_{\text{contact}}\),
\begin{align}
L_{\text{proximity}} &= \frac{1}{|V^I_h|} \sum_{v_h \in V^I_h} \max(0, d_{\text{ho}}(v_h) - \delta_{\text{contact}}).
\end{align}
\noindent where $d_{\text{ho}}(v_h)$ represents one-sided distances from \( V^I_h \) to \( V^I_o \).
We compute one-sidedly from hand to object because the hand mesh typically has more reliable geometry.
This avoids unstable gradients from noisy object predictions and encourages stable, plausible contact during optimization.
Both losses provide complementary guidance: the intersection loss enforces separation, while the proximity loss promotes firm
contact.
\par
\noindent \textbf{Final Phase 3 loss.} 
The total loss \(L_{\text{phase3}}\) sums the HOI 2D terms, the intersection and proximity losses, and softly reuses \(L_{\text{phase1}}\) and \(L_{\text{phase2}}\) with lower weights to avoid overriding new signals. 
At each step ($\tau_2 < t \le 1$) we optimize \((T_h,T_o)\) and steer \(\mathbf{v}_t\) for \(k_3\) iterations using \(L_{\text{phase3}}\), then update the latent via Eq.~\ref{eqn:diffusion_update}. 
At the end of this phase, we decode the final latent using the steered velocity field $\perturbedVelobj$, extract the object mesh, and pose both hand and object with the optimized $T_h$ and $ T_o$ to obtain the final HOI reconstruction \(\hoiM\).

%% file: fig/overview.tex
\begin{figure*}
    \centering
    \includegraphics[width=1.0\linewidth]{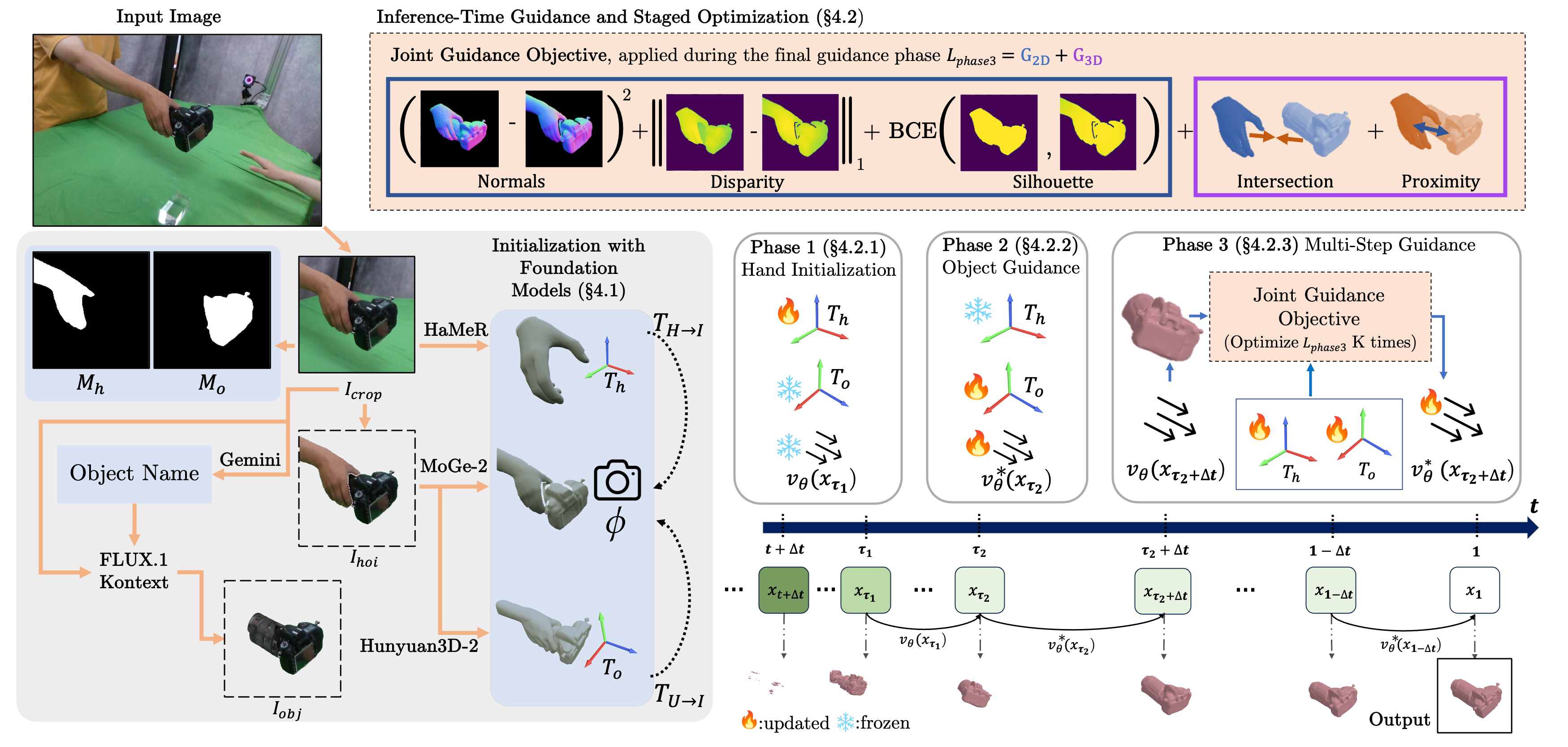}
    \caption{
\textbf{Overview of FollowMyHold}.
Given a single RGB frame, we (1) isolate the interaction region and derive binary hand/object masks with LangSAM and WiLoR's hand detector; (2) inpaint the occluded object appearance using FLUX.1 Kontext + Gemini (§4.1).
Next, we obtain three complementary 3D cues: a HaMeR hand mesh, a MoGe-2 partial point cloud (with camera pose $\phi$), and a coarse Hunyuan3D-2 HOI mesh. 
%
A two-step ICP registers all cues into a common image-aligned frame.
Finally, we perform inference-time guidance with a staged optimization (§4.2): Phase~1 optimizes the hand transform $T_h$; Phase~2 optimizes the object transform $T_o$ and guides the velocity field; Phase~3 jointly refines $(T_h,T_o)$ while guiding with pixel-aligned 2D losses ($G_{2\text{D}}$: normals, disparity, silhouette) and 3D constraints ($G_{3\text{D}}$: intersection, proximity).
The right bottom row shows progressive object refinement over diffusion steps.
}
\vspace{-10pt}
\label{fig:overview}
\end{figure*}

%% file: tab/algorithm.tex
\makeatletter
\algnewcommand{\LineComment}[1]{\Statex \hskip\ALG@thistlm \(\triangleright\) #1}
\makeatother

\begin{algorithm} 
\small
\caption{Inference-time Multi-step Optimization}
\label{algo:opt}
\begin{algorithmic}[1] 
\Require $\mathcal{R}_{S}$: Supervision maps \
\Require $\mathbf{w_{opt}}$: geometric optimization guidance strengths \
\Require DifferentiableRenderer\
\State $x_T \sim \mathcal{N}(0, I)$  \Comment{Sample noisy latent vector}
\ForAll{$t$ from $0$ to $1$}
\State $\mathbf{v}_t \leftarrow \mathcal{E}(x_t, t, c_{\text{obj}})$ 
\If {$t > \tau_2$} \Comment{Geometric guidance in Phase 3} \
\State $\mathbf{Z}_{t} = [\mathbf{v}_{t}, T_{o,t}, T_{h,t}]$  \Comment{Define optimization variables}
\For{$k = 1$ to $K$} \Comment{Multiple gradient descent steps}
\State $\hat{\mathbf{x}}^{k}_{1} = \mathbf{x}_t + (1 - \sigma_t) \cdot \mathbf{v}^{k}_{t} $
\State $\mathcal{M}^{U}_{o} = \mathrm{FlexiCubes}(\mathcal{D}(\hat{\mathbf{x}}^{k}_{1}, \mathbf{q}))$
\State $\mathcal{M}^{I}_{o} = T^{k}_{o,t} T_{U \rightarrow I}\mathcal{M}^{U}_{o}$, $\mathcal{M}^{I}_{h} = T^{k}_{h,t} T_{H \rightarrow I}\mathcal{M}^{H}_{h}$
\State $\mathcal{R}^{k}_{t} \leftarrow \text{DifferentiableRenderer}(\mathcal{M}^{I}_{o}\cup\mathcal{M}^{I}_{h})$ 
\LineComment{Compute 2D and 3D geometric losses}
\State ${G}_{2D} \leftarrow \mathcal{L}_{2D}(\mathcal{R}^{k}_{t}, \mathcal{R}_{S})$, ${G}_{3D} \leftarrow \mathcal{L}_{3D}(\mathcal{M}^{I}_{o}, 
\mathcal{M}^{I}_{h})$ 
\LineComment{Gradient-based updates}
\State $\mathbf{Z}^{k}_{t} \leftarrow \mathbf{Z}^{k}_{t} - \mathbf{w_{opt}}\nabla_{\mathbf{Z}^{k}_{t}} ({G}_{2D}(\mathbf{Z}^{k}_{t})+G_{3D}(\mathbf{Z}^{k}_{t}))$ 
\EndFor
\State $\mathbf{v}^{*}_{t} = \mathbf{v}^{K}_{t}$  \Comment{Update $\mathbf{v}$ with the steered velocity field}
\EndIf
\State $\mathbf{x}_t \leftarrow \mathbf{x}_t + \mathbf{v}^{*}_{t}$ \Comment{Rectified Flow step}
\EndFor
\LineComment{Obtain final hand and object meshes}
\State $\mathcal{M}^{I}_{o} = T_{o,1} T_{U \rightarrow I} \mathrm{FlexiCubes}(\mathcal{D}({\mathbf{x}}_1, q))$
\State $\mathcal{M}^{I}_{h} = T_{h,1} T_{H \rightarrow I}\mathcal{M}^{I}_{h,1}$
\end{algorithmic}
\end{algorithm}

%% file: sec/5_experiments.tex
\section{Experimental Results}
\label{sec:experiments}
We evaluate our approach on three publicly available datasets: OakInk \cite{yang2022oakink}, Arctic \cite{fan2023arctic}, DexYCB \cite{chao2021dexycb}.
All datasets include human grasps annotated with 3D hand-object poses, shapes, and meshes.
OakInk contains 100 rigid objects, Arctic 11 articulated objects, and DexYCB 20 YCB objects.
We randomly sample 1000 images from each dataset as our test sets.
Our model is not additionally trained as we use pretrained foundation models.
\input{tab/qualitative}
\input{fig/comparison_w_hort}
\input{fig/comparison_wo_hort}

\parag{Comparison Baselines.}
We compare against state-of-the-art hand–object interaction methods in two groups: deterministic feed-forward (FF) and generative (Gen). 
For FF methods, following~\cite{liu2024easyhoi}, we include IHOI~\cite{ye2022ihoi}, AlignSDF~\cite{alignsdf} and gSDF~\cite{chen2023gsdf}.
We also include HORT~\cite{chen2025hort}, which predicts point clouds; for fair comparison, we convert them to meshes using alpha shapes~\cite{Edelsbrunner1983_alphashapes} as in their Supplementary Sec.~B.1.
Our primary focus, however, is on Gen models, where EasyHOI currently sets the state-of-the-art by leveraging 2D and 3D foundation models for hand–object reconstruction.
Since AlignSDF, gSDF, and HORT were trained on DexYCB, we exclude them from DexYCB evaluation for fairness.
Video-based methods are excluded since our approach is single-frame.
%

\parag{Performance Measures.}
We assess three main aspects of the hand-object interaction reconstruction task.
%
Following \cite{chen2022alignsdf}, we report object accuracy via Chamfer Distance (CD) on $30$K point samples and F-scores at 5mm/10mm (F5/F10).
%
%
All object metrics (CD, F5, F10) capture both object shape and hand–object relative pose errors.
For grasp plausibility, we measure hand–object Intersection Volume (I.V.)~\cite{liu2024easyhoi}.
%
These accuracy metrics are computed only on the successfully reconstructed cases, excluding those where no output is provided, e.g., due to missing segmentation or failing optimization.
To assess how robust a method is, we compute the Reconstruction Rate (R.R), \textit{i.e.} the fraction of samples over the whole test set that the method produces an output for, regardless of quality.
More details regarding evaluation are in~\cref{supp:eval_details} of the supplemental material.
\par
\parag{Implementation Details.}
All experiments are conducted on an NVIDIA Tesla H100 NVL GPU. 
Our framework is implemented in PyTorch and integrated with the Hunyuan3D-2 pretrained diffusion backbone.
We use the Adam optimizer to update the hand and object transforms and to steer the velocity field.
As we optimize with respect to a single input view, the batch size is 1, and we use a total of 20 diffusion inference steps, with hand-only optimization conducted at step 9 and gradient-based guidance applied from step 10 onward.
The object guidance scale is fixed at $5.0$.
The remaining hyperparameters for each phase can be found in~\cref{supp:implementation} of the supplemental material.
Our method approximately takes $6.03$ minutes per sample.
\input{fig/phase_ablation}
\subsection{Comparison}
We report the quantitative comparison results of our method against the baselines in Tab. \ref{tab:qualitative}.
\textbf{Accuracy:} 
FollowMyHold achieves the lowest Chamfer Distance across all datasets, indicating superior \emph{global} alignment of reconstructed objects.
On OakInk, HORT achieves higher F-scores despite worse CD.
This indicates that HORT reconstructions are coarser than us (higher CD) while managing to not fail catastrophically on the challenging cases (higher F5/F10) as also indicated by their high reconstruction rate, due to its feed-forward design.
On Arctic, FollowMyHold generalizes better. 
HORT fails more frequently on thin, fine-grained, or larger tools of Arctic dataset, while FollowMyHold maintains coherent geometry and lower CD.
Furthermore, our approach maintains a relatively low intersection volume, significantly outperforming EasyHOI and HORT.
Although gSDF yields lower intersection volumes, it compromises significantly on object reconstruction accuracy and robustness.
\noindent \textbf{Robustness:} 
It is noteworthy that FollowMyHold achieves successful reconstruction at a significantly higher rate than EasyHOI ($87\%$ vs $39\%$ on OakInk), i.e. succeeds on more challenging cases while still being more accurate overall. 
In fact, our RR is comparable to feed-forward methods that rarely fail except under poor detections/segmentations.
This demonstrates the strength of our optimization-in-the-loop guidance, combining accuracy with high success rates among generative methods like EasyHOI.
\par
Figures~\ref{fig:comparison_w_hort} and~\ref{fig:comparison_wo_hort} show that our method accurately reconstructs object geometries and plausible hand-object interactions, surpassing EasyHOI, HORT, and IHOI.
%
%
EasyHOI struggles with scale consistency and object completeness, whereas IHOI, while better in grasp accuracy compared to EasyHOI, frequently fails in object reconstruction.
HORT struggles on rare, fine-grained geometries (e.g., Fig.~\ref{fig:comparison_w_hort}, second row, fourth column).
Point-cloud meshing via alpha-shapes (HORT Supplementary Sec.~B.1) can also break slender parts.
%
%
Further qualitative results are provided in~\cref{supp:qualitative} of the supplementary.
We also demonstrate our results on in-the-wild images, collected from the internet and our daily lives in Fig.~\ref{fig:teaser}, which shows FollowMyHold's robustness and accuracy in real-world examples.
\subsection{Ablation Study}
We analyze the effects of our staged design and influence of using guidance (Fig.~\ref{fig:phase_ablation} and Tab.~\ref{tab:phase_ablation}).
Our ablation set consists of $100$ random samples from OakInk dataset.
Our full method achieves 1.70cm$^2$ CD.
Without guidance (generate object first, then optimize transforms), CD rises to 3.70cm$^2$, and confirms the importance of guidance for object refinement.
Without Phase~1 (no hand-only optimization) CD degrades to 5.67cm$^2$, confirming the hand as a spatial anchor.
Without Phase~2 (no object-only optimization) CD increases to 3.53cm$^2$ and indicates that object-before-joint refinement is necessary.
Swapping the order of Phase~1 and Phase~2 results in 3.18cm$^2$, which demonstrates the importance of the phase order.
When we perform hand-object joint optimization from the start, CD increases to 5.51cm$^2$ and highlights the importance of having phases of optimization.
Additionally, to show that our main performance increase compared to EasyHOI comes from our method design rather than Hunyuan3D-2, we replace EasyHOI’s mesh generator with Hunyuan3D-2, which only slightly improves CD (5.23~$\rightarrow$~5.13cm$^2$). 
Our method still achieves a CD nearly $3\times$ lower.
Supplemental material includes the corresponding details in~\cref{supp:ablation_easyhoi} and our ablation on loss terms in~\cref{supp:ablation_lossterms}.
\input{tab/phase_ablation}
%
%
%

%% file: tab/qualitative.tex
\begin{table*}
  \centering
  \caption{
  \textbf{Quantitative evaluation for HOI reconstruction.} 
  We do not evaluate AlignSDF, gSDF, and HORT on DexYCB as they are trained on it.
  F5 and F10 measure the F-scores of reconstructed object points within 5mm and 10mm of the ground-truth (GT) object, respectively. 
  C.D.\ refers to the Chamfer Distance (in cm$^2$) between the reconstructed object and GT object.
  I.V.\ refers to Intersection Volume (in cm$^3$) between the hand mesh and the object mesh.
  R.R. refers to the Reconstruction Rate, defined as the fraction of objects for which a method produces an output, regardless of quality.
  Methods are also categorized as feed-forward (FF) or generative (Gen).
  }
   \label{tab:qualitative}
   \resizebox{\linewidth}{!}{
  \begin{tabular}{c l
                  *{5}{c}
                  *{5}{c}
                  *{5}{c}}
    \toprule
      & Method
      & \multicolumn{5}{c}{OakInk}
      & \multicolumn{5}{c}{Arctic}
      & \multicolumn{5}{c}{DexYCB} \\
    \cmidrule(lr){3-7} \cmidrule(lr){8-12} \cmidrule(lr){13-17}
      & 
      & F5 $\uparrow$ & F10 $\uparrow$ & C.D.\ $\downarrow$ & I.V.\ $\downarrow$ & R.R.\ $\uparrow$
      & F5 $\uparrow$ & F10 $\uparrow$ & C.D.\ $\downarrow$ & I.V.\ $\downarrow$ & R.R.\ $\uparrow$
      & F5 $\uparrow$ & F10 $\uparrow$ & C.D.\ $\downarrow$ & I.V.\ $\downarrow$ & R.R.\ $\uparrow$ \\
    \midrule
    \multirow{3}{*}{%
  \rotatebox{90}{FF}
} 
   & IHOI
      & 0.132 & 0.252 & 3.97 & 6.20 & 0.82
      & 0.079 & 0.148 & 11.9 & \uline{4.27}  & 0.63
      & \uline{0.108} & \uline{0.211} & \uline{4.87} & \textbf{6.41} & \uline{0.49}  \\
    & AlignSDF
      & 0.060 & 0.119 & 7.75  & 12.5 & 0.94
      & 0.042 & 0.083 & 17.7 & 15.6 & 0.86
      & --    & --    & --    & --   & -- \\
   & gSDF
      & 0.047 & 0.093 & 8.17 & \textbf{0.62} & \uline{0.87}
      & 0.036 & 0.070 & 18.9 & \textbf{1.23}  & 0.79
      & --    & --    & --    & --   & -- \\
    & HORT
      & \textbf{0.319} & \textbf{0.508} & \uline{2.22} & 16.6 & \textbf{0.99}
      & \uline{0.101} & \uline{0.190} & 11.0 & 30.8 & \uline{0.88}
      & -- &  -- & -- & --  & -- \\
    \midrule
    \multirow{2}{*}{%
  \rotatebox{90}{Gen}
}
   & EasyHOI
      & 0.109 & 0.210 & 4.62 & 21.31 & 0.39 
      & 0.079 & 0.145 & \uline{10.9} & 18.3 & 0.36
      & 0.090 & 0.176 & 6.26 & 19.13 & 0.30  \\
   & \textbf{Ours}
      & \uline{0.179} & \uline{0.322} & \textbf{1.80} & \uline{5.96} & \uline{0.87}
      & \textbf{0.160} & \textbf{0.288} & \textbf{2.57} & 5.08 & \textbf{0.92}
      & \textbf{0.158} & \textbf{0.300} & \textbf{2.04} & \uline{7.02} & \textbf{0.58}  \\
    \bottomrule
  \end{tabular}
  }
\end{table*}

%% file: fig/comparison_w_hort.tex
\begin{figure*}
    \centering
    \includegraphics[width=1.0\linewidth]{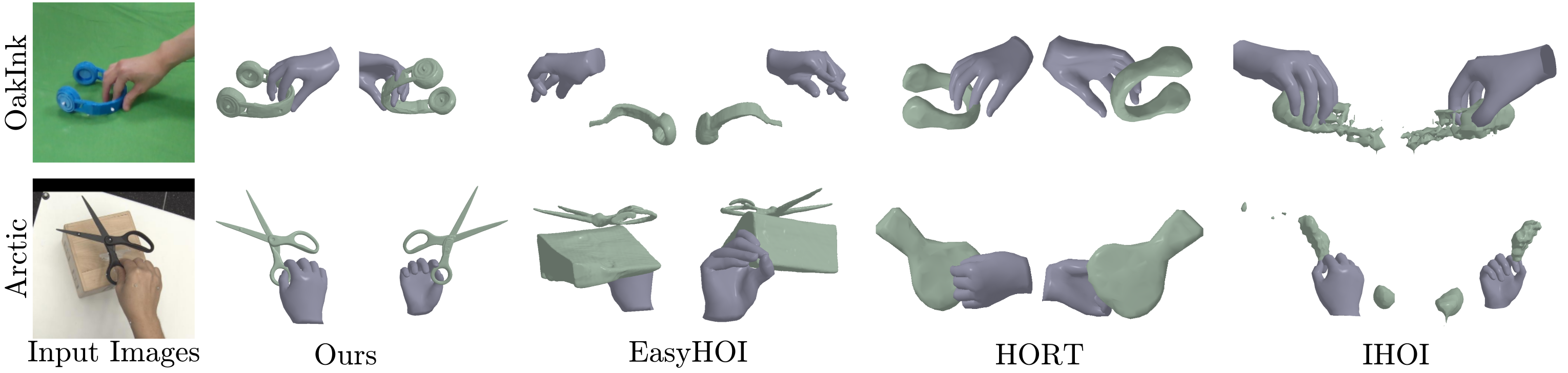}
    \caption{
    \textbf{Qualitative comparison of 3D HOI reconstructions on OakInk and Arctic datasets.} 
    From left to right: input RGB frame; reconstructions produced by our method, by EasyHOI, by HORT, and by IHOI.
    We evaluate and visualize HORT results after converting the point clouds into meshes following the procedure mentioned in the HORT paper, Supplementary Section B.1.
    Our approach yields more accurate object geometry with more plausible hand-object interactions.
    }
    \label{fig:comparison_w_hort}
\end{figure*}

%% file: fig/comparison_wo_hort.tex
\begin{figure}
    \centering
    \includegraphics[width=1.0\linewidth]{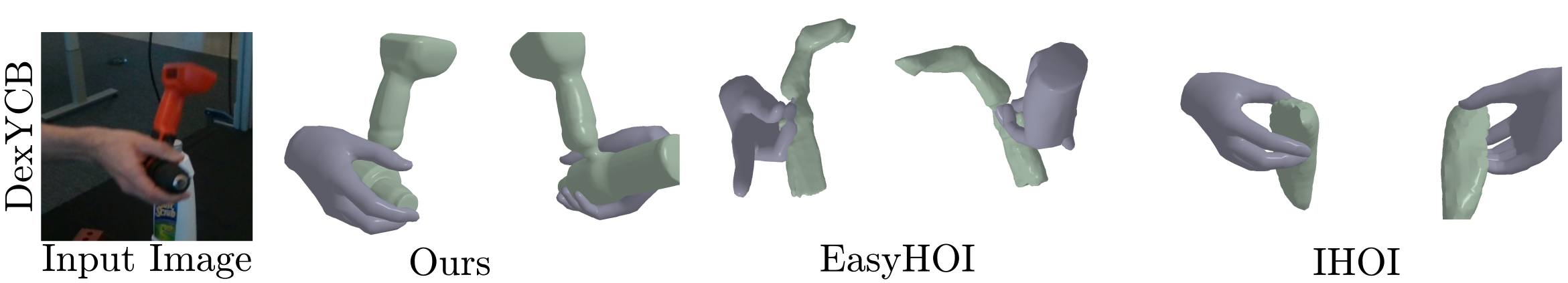}
    \caption{
    \textbf{Qualitative comparison of 3D HOI reconstructions on DexYCB dataset.} 
    From left to right: input RGB frame; reconstructions produced by our method, by EasyHOI, and by IHOI.
    Since HORT is trained on DexYCB, we do not include it to the comparison.
    Our method yields more accurate geometry and plausible interactions.
    }
    \label{fig:comparison_wo_hort}
\end{figure}

%

%% file: fig/phase_ablation.tex
\begin{figure*}
    \centering
    \includegraphics[width=1.0\linewidth]{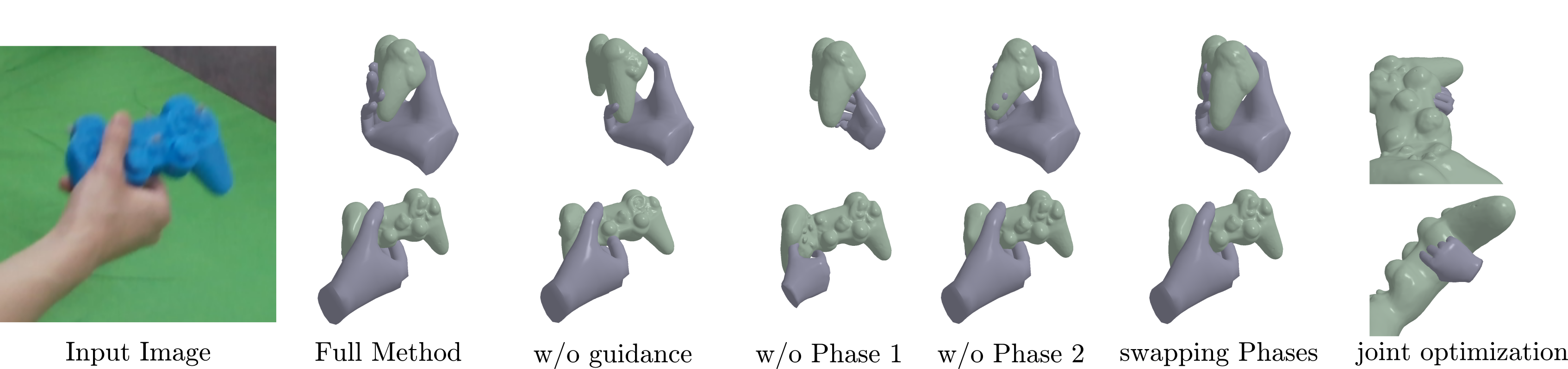}
    \caption{
    \textbf{Ablation of the phased optimization.} 
    We visualize the ablation study on our staged optimization design.
    We compare HOI reconstruction quality when removing inference-time guidance, dropping Phase 1 or Phase 2, swapping their order, or performing joint optimization. 
    The full method yields the most accurate grasp and object.
    The bottom row shows the camera viewpoint, while the top row provides an alternative view to highlight differences.
    }
    \label{fig:phase_ablation}
    \vspace{-10pt}
\end{figure*}

%% file: tab/phase_ablation.tex
\begin{table}
\small
  \centering
  \caption{
  \textbf{Phase ablation.}
  Ablation study for each method phase on the ablation set. 
  }
  \label{tab:phase_ablation}
  \begin{tabular}{lcc}
    \toprule
    Method & C.D. $\downarrow$ & I.V. $\downarrow$ \\
    \midrule
    w/o inference-time guidance & 3.70 & 5.83 \\
    w/o Phase 1 & 5.67 & 7.66 \\
    w/o Phase 2 & 3.53 & 4.82 \\
    swapping Phases & 3.18 & 5.19 \\
    joint optimization & 5.51 & 11.7 \\
    \midrule
    Full Method & \textbf{1.70} & \textbf{4.03} \\
    \bottomrule
  \end{tabular}
  \vspace{-10pt}
\end{table}

%% file: sec/6_conclusion.tex
%
%

\section{Limitations and Future Work} 
%

%
%
%
%
%
%
Our inference-time guidance trades accuracy and robustness for compute: each step includes inner-loop gradient updates with backpropagation through the latent decoder, increasing runtime.
Our method also assumes reliable upstream segmentation and inpainting, however artifacts at this stage can propagate into reconstruction. 
Another limitation lies in reconstructing thin objects where supervision signals are weaker.
Examples are shown in~\cref{supp:limitations} of the supplemental material.
\par
Future work includes more robust fusion of 2D and 3D signals with learned interaction priors for uncertain regions, integrating our geometric objectives into diffusion training to eliminate the guidance loop, and extending to video with temporal consistency (e.g., optical flow) for stable AR/VR and telepresence reconstructions.
\section{Conclusion}
%
%
We introduced \textbf{FollowMyHold}, a single-image HOI reconstruction method that guides a pretrained 3D rectified-flow model at inference using an optimization-in-the-loop design. 
We show that the proposed inference-time guidance from pixel-aligned 2D cues (normal and disparity alignment, silhouette consistency) together with 3D interaction constraints (intersection, proximity), applied within a staged optimization (hand~$\rightarrow$~object~$\rightarrow$~joint), produce physically plausible and globally consistent reconstructions.
Interestingly, this strategy significantly improves the reconstruction robustness, in addition to the improved object reconstruction quality.

%% file: sec/7_supplementary.tex
\clearpage
\setcounter{page}{1}
\maketitlesupplementary
\section*{Overview}
This supplementary material provides additional details and results to complement the main paper. 
Evaluation protocols are detailed in \cref{supp:eval_details}.
\cref{supp:method_details} presents further methodological details, such as the derivation of the clean sample estimate and the full per-phase loss functions. 
In \cref{supp:implementation}, we describe the implementation setup, including hyper-parameters and learning rates for reproducibility. 
In \cref{supp:limitations}, we elaborate and illustrate the limitations of our method.
\cref{supp:ablations} extends the ablation study by analyzing the effect of individual loss terms (\cref{supp:ablation_lossterms}) and the EasyHOI ablation (\cref{supp:ablation_easyhoi}). 
We further compare our reconstructions with HORT on the Arctic~\cite{fan2023arctic} dataset (\cref{supp:hort_vs_ours}), and illustrate the coarseness of Hunyuan3D-2~\cite{hunyuan} outputs in \cref{supp:coarse_hunyuan}. 
Intermediate results from MoGe-2~\cite{moge} and Hunyuan3D-2 are shown in \cref{supp:intermediate} to highlight their partial/coarse nature compared to our reconstructions. 
Finally, \cref{supp:qualitative} provides extensive qualitative results.
\section{Evaluation Details}
\label{supp:eval_details}
We evaluate object reconstruction quality using Chamfer Distance (CD) between $30$K sampled points from the prediction and the reference mesh. 
Following \cite{chen2022alignsdf}, reconstructed hand meshes are first aligned to the ground-truth hand mesh in scale, rotation, and translation via ICP, and the same transformation is applied to the reconstructed object. 
All object metrics (CD, F5, F10) are therefore computed on these hand-aligned meshes, jointly capturing object geometry and hand–object relative pose errors. 
For grasp plausibility, we follow \cite{liu2024easyhoi} and compute the hand–object intersection volume (I.V.) using the \texttt{trimesh} library with a voxel size of $0.5$cm.
\section{Method Details}
\label{supp:method_details}
\subsection{Derivation of Clean Sample Estimate}
\label{supp:clean_sample_estimate}
We build on the affine path formulations introduced by Lipman et al.~\cite{lipman2024flow} (Eq.~4.54), which provides equivalent parameterizations of affine paths. 
Following Hunyuan3D-2, we adopt the convention where $\mathbf{x}_1$ denotes the target and $\mathbf{x}_0$ the noise. 
Relative to Lipman et al., this requires swapping the subscripts $0$ and $1$, yielding
\begin{equation}
    \mathbf{x}_1 = \frac{\mathbf{x}_t - \alpha_t \mathbf{x}_0}{\sigma_t},
\label{supp_eq:flow4.54}
\end{equation}
with a conditional optimal transport schedule defined by $\alpha_t = t$ and $\sigma_t = 1-t$.  
\par
\noindent In Hunyuan3D-2, the velocity field is defined as
\begin{equation}
    \velocity = \mathbf{x}_1 - \mathbf{x}_0.
\label{supp_eq:hunyuan_velocity}
\end{equation}
Substituting $\mathbf{x}_0$ from \cref{supp_eq:hunyuan_velocity} into \cref{supp_eq:flow4.54} gives
\begin{equation}
\begin{aligned}
    \mathbf{x}_1 &= \frac{\mathbf{x}_t - \alpha_t (\mathbf{x}_1 - \velocity_t)}{\sigma_t}, \\
    \sigma_t \mathbf{x}_1 &= \mathbf{x}_t - (1-\sigma_t)(\mathbf{x}_1 - \velocity_t), \\
    \mathbf{x}_1 &= \mathbf{x}_t + (1-\sigma_t)\velocity_t,
\end{aligned}
\end{equation}
which yields our clean target estimate. 
With this formulation, we can recover the clean target estimate $\hat{\mathbf{x}}_1$ at any step to perform guidance.
\subsection{Per-phase Losses}
\label{supp:full_objectives}
The optimization objectives for each phase are defined as follows.  
\paragraph{Phase 1 (hand optimization).} 
The objective for optimizing $T_h=(s_h, R_h, t_h)$ is
\begin{equation}
\begin{split}
L_{\mathrm{phase1}}(s_h, R_h, t_h) &=
  \lambda_1 L_{2d} +
  \lambda_2 L_{\mathrm{norm,h}} +
  \lambda_3 L_{\mathrm{disp,h}} \\
  &\quad +
  \lambda_4 L_{\mathrm{sil,h}} +
  \lambda_5 L_{\mathrm{reg,h}}.
\end{split}
\end{equation}
\paragraph{Phase 2 (object optimization).} 
The total loss for optimizing $T_o=(s_o, R_o, t_o)$ while guiding $\velocity$ is
\begin{equation}
\begin{split}
L_{\mathrm{phase2}} &=
  \alpha_1 L_{\text{norm,o}} +
  \alpha_2 L_{\text{disp,o}} +
  \alpha_3 L_{\text{sil,o}} +
  \alpha_4 L_{\text{reg,o}}.
\end{split}
\end{equation}
\paragraph{Phase 3 (joint HOI optimization).} 
To optimize $T_h, T_o$ and guide $\velocity$, we combine HOI-level 2D and 3D constraints with the Phase 1 and Phase 2 objectives:
\begin{equation}
\begin{split}
L_{\mathrm{phase3}} &= 
  \gamma_1 L_{\text{norm,hoi}} +
  \gamma_2 L_{\text{disp,hoi}} +
  \gamma_3 L_{\text{sil,hoi}} \\
  &\quad +
  \gamma_4 L_{\text{intersection}} +
  \gamma_5 L_{\text{proximity}} \\
  &\quad +
  \gamma_{6} L_{\mathrm{phase1}} +
  \gamma_{7} L_{\mathrm{phase2}}.
\end{split}
\end{equation}
\subsection{Proximity Loss Detail}
To encourage realistic hand-object contact while avoiding such excessive separation, we compute one-sided distances from \( V^I_h \) to \( V^I_o \) by finding the closes vertex $v_h$,
\begin{equation}
d_{\text{ho}}(v_h) = \min_{v_o \in V^I_o} \|v_h - v_o\|_2, \quad \forall v_h \in V^I_h.
\end{equation}
Proximity loss penalizes overly large gaps while still allowing small contact distances.
\section{Implementation Details}
\label{supp:implementation}
\paragraph{Hyper-parameters.}
For Phase 1 losses, we use 
$\lambda_1=0.01, 
\lambda_2=1, 
\lambda_3=10, 
\lambda_4=10, 
\lambda_5=0.01$.
In Phase 2, we set 
$\alpha_1=10,  
\alpha_2=10,  
\alpha_3=10,  
\alpha_4=0.001$.
For Phase 3, the loss weights are: $ \gamma_1=10,
  \gamma_2=10,
  \gamma_3=10,
\gamma_4=0.00001,
  \gamma_5=1,
\gamma_{6}=0.001,
  \gamma_{7}=0.001$.
We further use $\tau_1=0.526$ and $\tau_2=0.579$ in our optimization-in-the-loop guidance design.  
\paragraph{Phase 1 Learning Rates.} For optimizing hand transformations, we use a learning rate of $0.01$ for scale, $0.01$ for translation, and $0.5$ for rotation. 
Optimization is run for $200$ steps.  
\paragraph{Phase 2 Learning Rates.} For object transformations, we use $0.05$ for scale, $0.05$ for translation, and $0.01$ for rotation. 
The velocity field is optimized with a rate of $0.0001$. 
Optimization runs for $100$ steps.  
\paragraph{Phase 3 Learning Rates.} 
For hand transformations, we use $0.0001$ for scale, $0.0001$ for translation, and $0.01$ for rotation. 
For object transformations, we use $0.005$ for scale, $0.005$ for translation, and $0.01$ for rotation. 
The velocity field is guided with a rate of $0.01$. 
Joint refinement of hand and object is performed over $50$ steps.  
\section{Limitations}
\label{supp:limitations}
Our approach increases computational cost since each diffusion step involves inner-loop gradient updates with backpropagation through the latent decoder. 
This is also why the Initialization with Foundation Models (\cref{subsec:init}) takes $\approx 2$ minutes while Inference-Time Guidance and Staged Optimization (\cref{sec:guidance}) takes $\approx 4$ minutes.
Additionally, our method assumes reliable upstream segmentation and inpainting; errors in these stages may propagate to reconstruction, as shown in~\cref{fig:fails} second row, where incorrect inpainting leads to reconstructing the wrong object (two cans). 
Finally, reconstructing thin or fine-grained structures remains challenging, since both 2D and 3D guidance signals may be insufficient to accurately optimize such poses (e.g., glasses in the first row of~\cref{fig:fails}). 
\input{fig/failure_cases}
\section{Ablation Study}
\label{supp:ablations}
\subsection{Ablation on Loss Terms}
\label{supp:ablation_lossterms}
We systematically analyze the influence of individual loss components used during optimization and guidance in~\cref{tab:loss_ablation} and \cref{fig:loss_ablation}.
Removing any single loss component deteriorates object and grasp quality, and reconstruction rate.
Omitting the normal map loss (third column) leads to incorrect object scale and intersection with the hand, while removing the depth loss (fourth column) causes incorrect object scale, misalignment and additional intersection.
Without the silhouette loss (fifth column), optimization diverges, producing severe scale mismatch and mesh penetration.
Dropping the intersection penalty (sixth column) results in increased misalignment and hand-object intersection.
Finally, excluding the proximity loss (last column) increases hand-object distance and causing the grasp to vanish.
Using the full loss formulation (second column) yields physically plausible and well-aligned grasps.
Employing the full set of losses ensures robust optimization, achieving firm hand-object contacts and accurate geometry reconstructions.
\input{tab/loss_ablation}
\input{fig/loss_ablation}
\subsection{Ablation on EasyHOI}
\label{supp:ablation_easyhoi}
To verify that our superior performance over EasyHOI comes from our method design rather than the use of Hunyuan3D-2, we replace EasyHOI’s mesh generator InstantMesh~\cite{xu2024instantmesh} with Hunyuan3D-2, while keeping all other components of EasyHOI (segmentation, inpainting, optimization) unchanged. 
To ensure comparable initialization, Hunyuan3D-2 meshes are aligned to InstantMesh outputs via ICP before being passed to EasyHOI’s optimizer. 
As shown in \cref{tab:easyhoi_ablation} and \cref{fig:easyhoi_ablation}, this variant still underperforms our method by a large margin, confirming that the gain comes primarily from our optimization framework.
\input{tab/easyhoi_ablation}
\input{fig/easyhoi_ablation}

\section{Comparisons}
\subsection{FollowMyHold vs HORT}
\label{supp:hort_vs_ours}
\input{fig/hort_ours}
We visually compare our reconstructions with HORT at \cref{fig:hort_vs_ours_arctic}. 
Since HORT is not trained on datasets containing large, detailed objects, it struggles to reconstruct them, highlighting the advantage of our approach, which leverages large pretrained models for generalization without task-specific training. 
\par
In the first row, the input contains a large box: our method reconstructs it accurately, while HORT fails, leading to severe hand–object intersections. 
In the second row, with a fine-grained object, our reconstruction preserves the details and the grasp, whereas HORT does not. 
In the third row, which includes a large, fine-grained object, our method produces a plausible reconstruction (except for the lower box attached due to input ambiguity), while HORT yields a coarse mesh that causes hand-object penetration.  
For fairness, we converted HORT’s point cloud outputs to meshes using the alpha-shapes algorithm, following HORT Supplementary Sec.~B.1. 
This conversion further increases the coarseness of the HORT meshes.
\subsection{Coarse HOI Meshes from Hunyuan3D-2}
\label{supp:coarse_hunyuan}
\input{fig/coarse_hunyuan}
We visualize HOI meshes reconstructed by Hunyuan3D-2 from a single HOI input image in \cref{fig:coarse_hunyuan}. 
Although these meshes are coarse, they still provide useful cues about the relative positioning of the hand and object, which we leverage in \cref{subsubsec:icp} for ICP alignment.
\subsection{Comparison with Intermediate Results}
\label{supp:intermediate}
We evaluate the quality of hand-object interaction reconstructions produced by MoGe-2 and Hunyuan3D-2, and compare them to our final result in Fig.~\ref{fig:intermediate}.
MoGe-2 generates only partial object geometry, with large missing regions such as the backside.
Hunyuan3D-2 produces a complete object mesh, but the hand is poorly reconstructed and appears fused with the object, lacking separation and plausible contact, also illustrated in \cref{supp:coarse_hunyuan}.
In contrast, our method leverages these intermediate signals, MoGe-2’s image-aligned partial geometry and Hunyuan3D-2’s object geometry, and integrates them into a physically plausible, full hand-object reconstruction with clear separation and contact, as shown in the second column in \cref{fig:intermediate}.
\input{fig/intermediate}
\section{Qualitative Results}
\label{supp:qualitative}
We present additional qualitative results on OakInk (Figures~\ref{fig:oakink_1}–\ref{fig:oakink_8}), Arctic (Figures~\ref{fig:arctic_1}–\ref{fig:arctic_4}), and DexYCB (Figures~\ref{fig:dexycb_1} and \ref{fig:dexycb_2}). 
The first column, labeled \textit{Input Images}, shows hand–object interaction crops obtained after the preprocessing step in \cref{subsubsec:preprocessing2d}.
\input{fig/more_qualitative/supp1}
\input{fig/more_qualitative/supp2}
\input{fig/more_qualitative/supp3}
\input{fig/more_qualitative/supp4}
\input{fig/more_qualitative/supp5}
\input{fig/more_qualitative/supp6}
\input{fig/more_qualitative/supp7}
\input{fig/more_qualitative/supp8}
\input{fig/more_qualitative/supp9}
\input{fig/more_qualitative/supp10}
\input{fig/more_qualitative/supp11}
\input{fig/more_qualitative/supp12}
\input{fig/more_qualitative/supp13}
\input{fig/more_qualitative/supp14}

%% file: fig/failure_cases.tex
\begin{figure}
    \centering
    \includegraphics[width=1.0\linewidth]{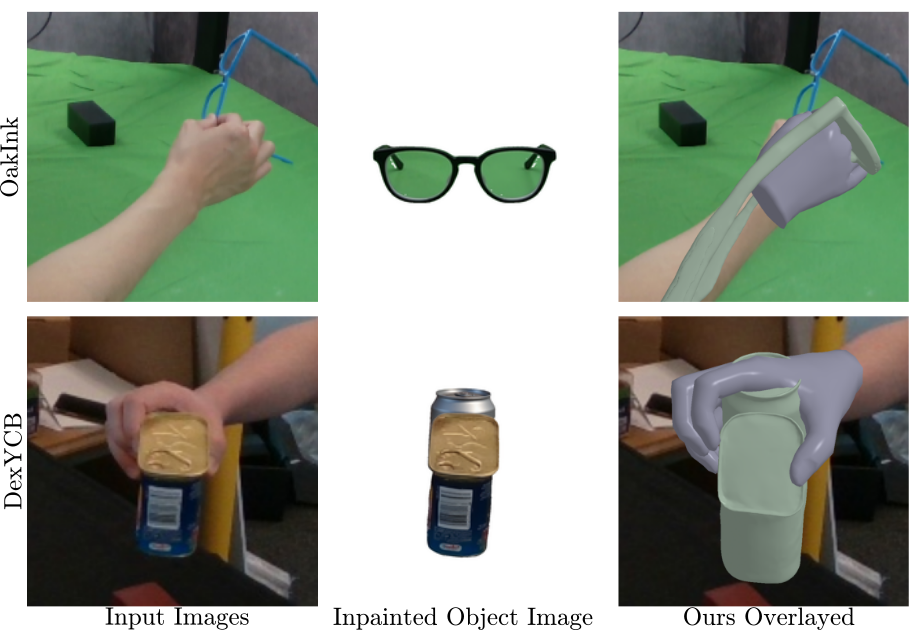}
    \caption{\textbf{Failure cases.}
    First column: input images; second: inpainted and masked object appearances; third: our reconstructions overlaid on the input. 
    Failure modes include difficulties with thin structures such as glasses in the first row, where 2D and 3D signals are insufficient for accurate pose optimization, and errors from incorrect inpainting (second row), which lead the diffusion model to generate the wrong object (two cans).}
    \label{fig:fails}
\end{figure}

%% file: tab/loss_ablation.tex
\begin{table}
  \centering
   \caption{\textbf{Loss ablation.} 
   Effect of removing individual loss terms from the objective on object reconstruction quality, evaluated on our ablation set.}
  \label{tab:loss_ablation}
  \begin{tabular}{lccc}
    \toprule
    Method & C.D. $\downarrow$ & I.V.$\downarrow$ & R.R.$\uparrow$ \\
    \midrule
    w/o $L_{\text{normal}}$   & 2.10  & 6.00 & 0.84 \\
    w/o $L_{\text{depth}}$    & 2.13 & 4.84 & \textbf{0.87} \\
    w/o $L_{\text{silhouette}}$ & 4.29 & 7.79 & 0.78 \\
    w/o $L_{\text{intersection}}$ & 2.59 & 9.33 & 0.86\\
    w/o $L_{\text{proximity}}$   & 3.43 & \textbf{3.56} & 0.85 \\
    \midrule
    Full losses            & \textbf{1.70} & 4.03 & \textbf{0.87} \\
    \bottomrule
  \end{tabular}
\end{table}

%% file: fig/loss_ablation.tex
\begin{figure*}
    \centering
    \includegraphics[width=1.0\linewidth]{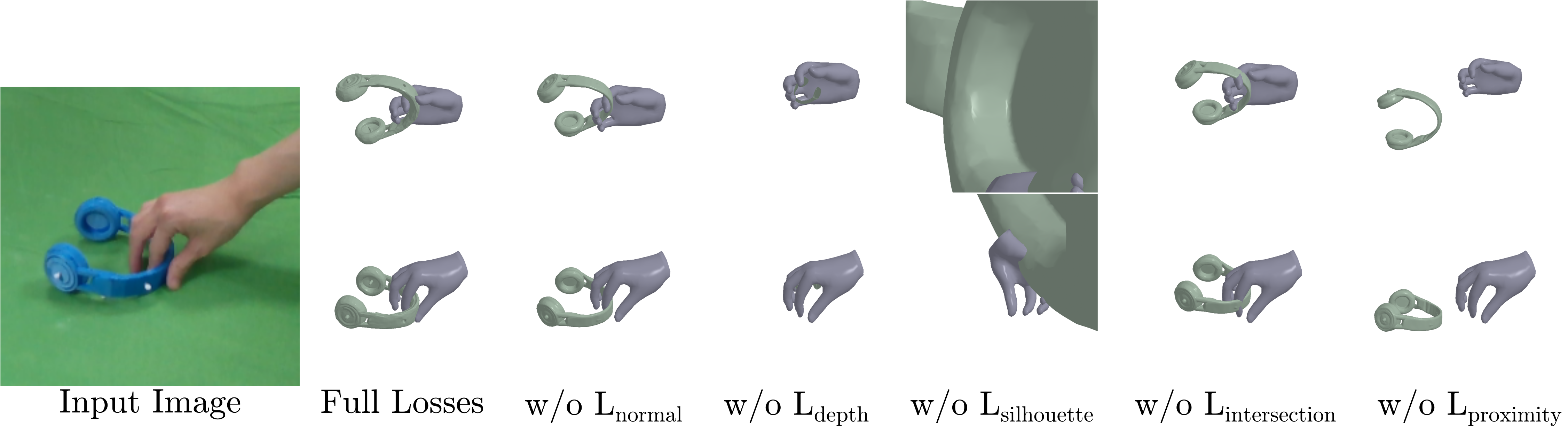}
    \caption{
    \textbf{Ablating the objective terms.} 
    For each loss term, we show the HOI reconstruction when it is removed from all optimization phases. 
    Bottom row shows the input viewpoint while the top row provides an alternative view to better illustrate the grasp. 
    Removing any objective term degrades interaction quality.
    Using the full loss formulation (second column) yields physically plausible and well-aligned grasps.
    }
    \label{fig:loss_ablation}
\end{figure*}

%% file: tab/easyhoi_ablation.tex
\begin{table}
  \centering
  \caption{
  \textbf{EasyHOI ablation.}
  Replacing InstantMesh with Hunyuan3D-2 in EasyHOI shows that our performance gains come from our optimization design rather than the choice of the object generator.
  }
  \label{tab:easyhoi_ablation}
  \begin{tabular}{lccc}
    \toprule
    Method & C.D. $\downarrow$ & I.V. $\downarrow$ & R.R. $\uparrow$ \\
    \midrule
    EasyHOI w/ InstantMesh & 5.23 & 22.4 & 0.44 \\
    EasyHOI w/ Hunyuan3D-2 & 5.13 & 22.0 & 0.39 \\
    \midrule
    Ours & \textbf{1.70} & \textbf{4.03} & \textbf{0.87} \\
    \bottomrule
  \end{tabular}
  \vspace{-10pt}
\end{table}

%% file: fig/easyhoi_ablation.tex
\begin{figure*}
    \centering
    \includegraphics[width=1.0\linewidth]{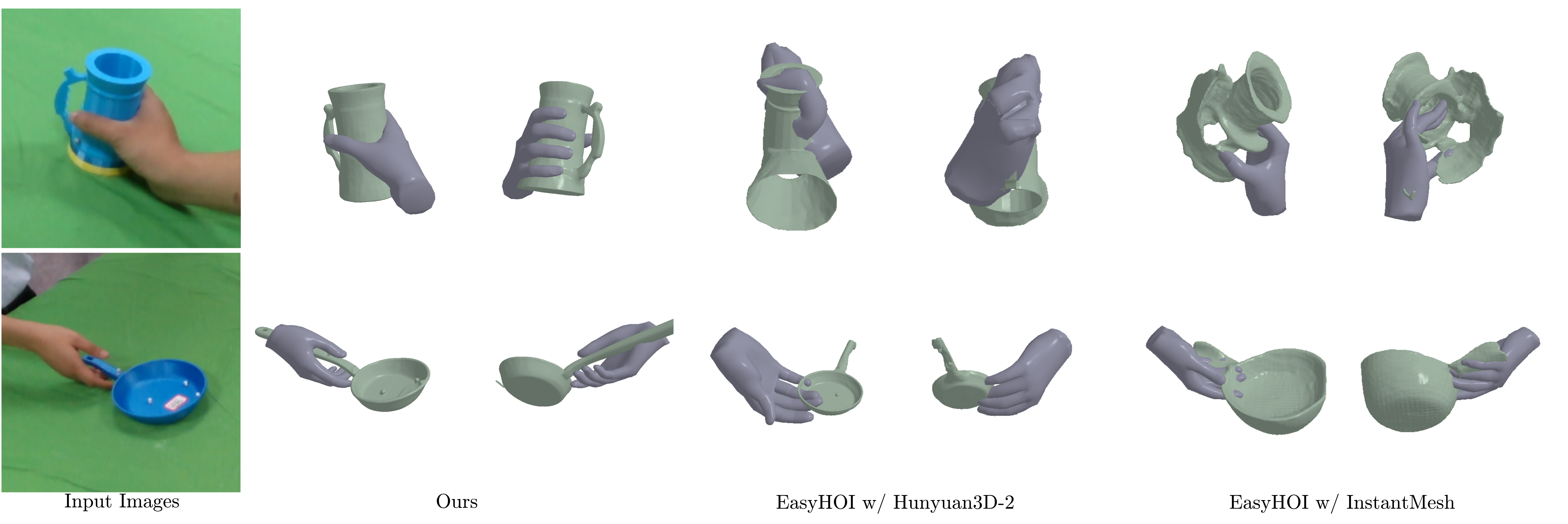}
    \caption{
    \textbf{Ablation of EasyHOI with mesh generator.} 
    We visualize the ablation on EasyHOI by exchanging its mesh generator InstantMesh~\cite{xu2024instantmesh} with the mesh generator we utilize, Hunyuan3D-2~\cite{hunyuan}.
    }
    \label{fig:easyhoi_ablation}
\end{figure*}

%% file: fig/hort_ours.tex
\begin{figure*}
    \centering
    \includegraphics[width=1.0\linewidth]{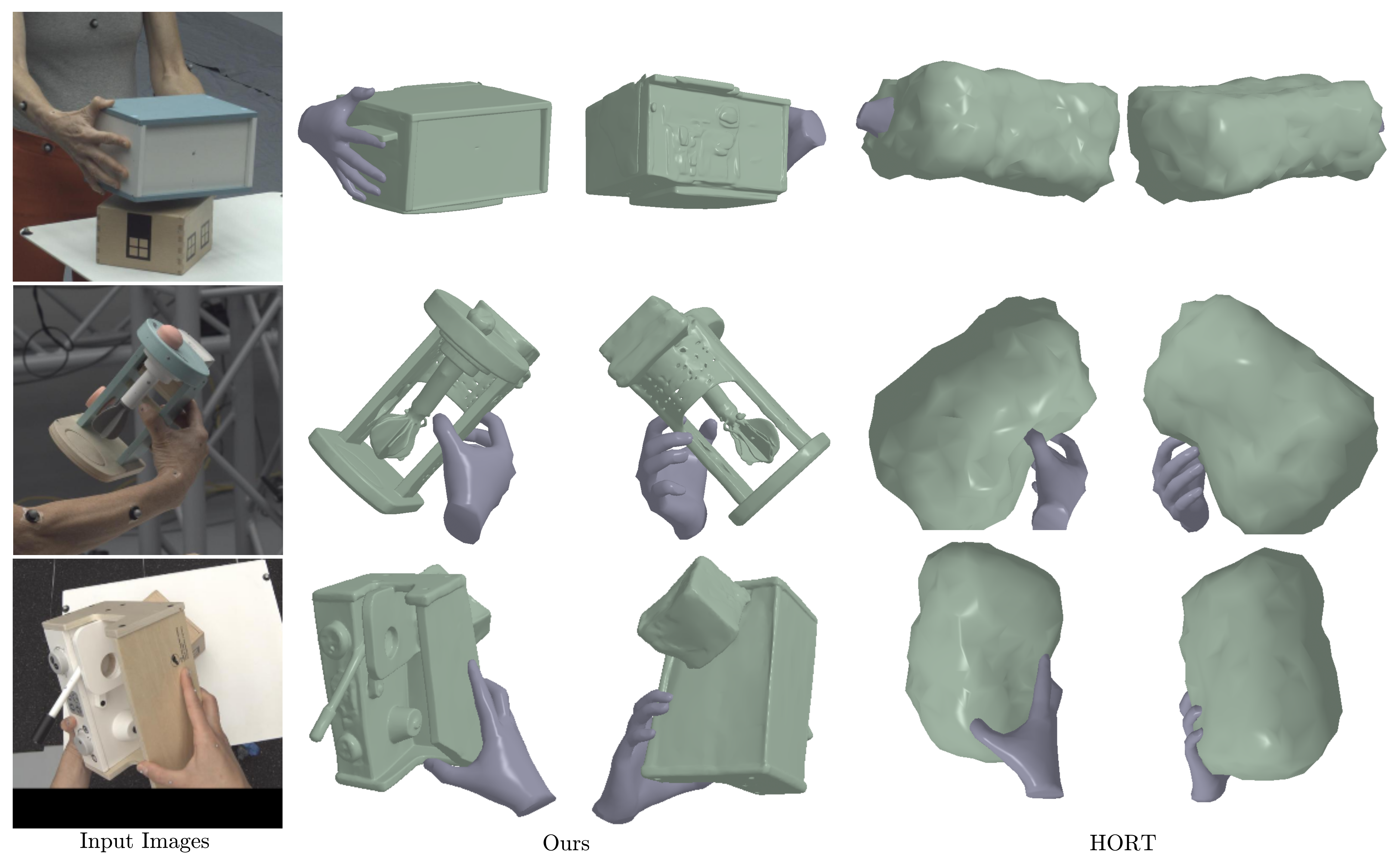}
    \caption{
    \textbf{Qualitative comparison between HORT and 3D HOI reconstructions Arctic dataset.} 
    From left to right: input RGB frame; reconstructions produced by our method, and by HORT.
    First row represents an input with a large object.
    Second row represents a fine-grained object.
    Third row represents a large, fine-grained object.
    As shown, our HOI reconstructions are much more accurate and detailed compared to HORT's reconstructions.
    }
    \label{fig:hort_vs_ours_arctic}
\end{figure*}

%% file: fig/coarse_hunyuan.tex
\begin{figure}
    \centering
    \includegraphics[width=1.0\linewidth]{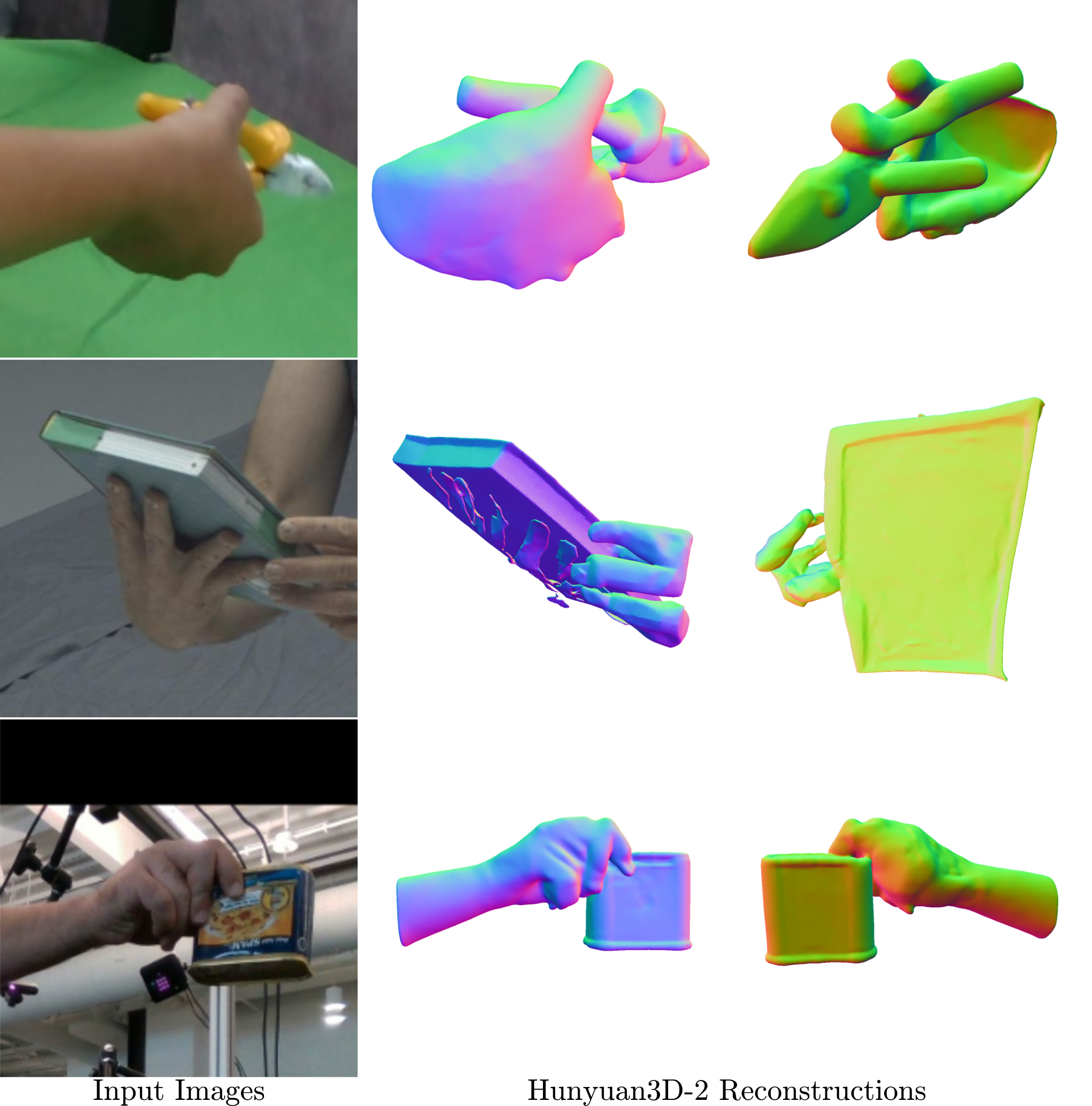}
    \caption{
    \textbf{Coarse HOI reconstructions of Hunyuan3D-2.} 
    From left to right: input RGB frame, normals of the reconstructed HOI mesh from the input view, and the same mesh from an alternative view. 
    The first, second, and third rows correspond to OakInk, Arctic, and DexYCB, respectively.
    }
    \label{fig:coarse_hunyuan}
\end{figure}

%% file: fig/intermediate.tex
\begin{figure}
    \centering
    \includegraphics[width=1.0\linewidth]{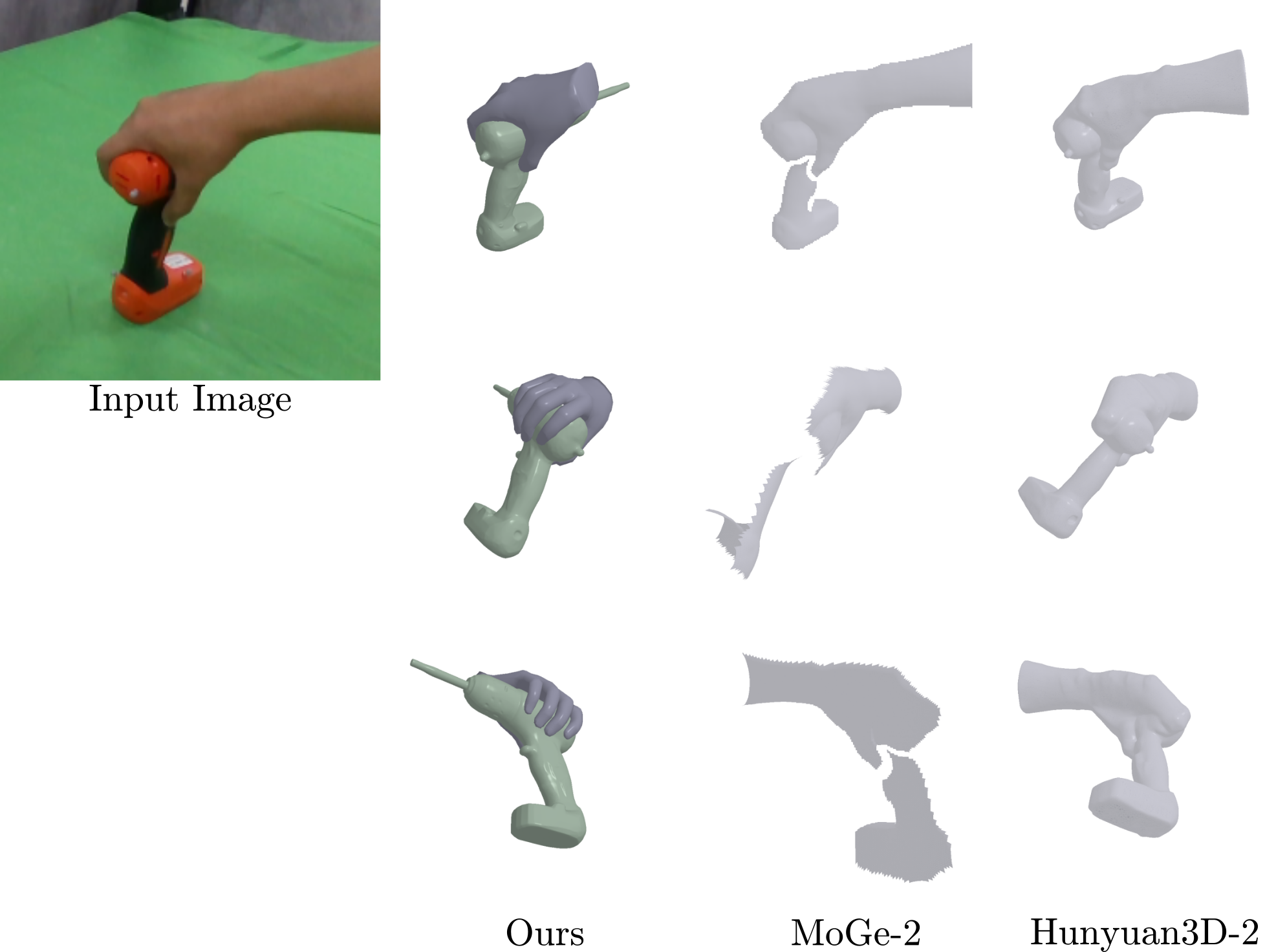}
    \caption{
    \textbf{Intermediate results.}
From top to bottom, we show front, side, and back views. 
The second column presents our full hand-object reconstruction on an OakInk example.
The third column shows renderings from MoGe-2, which outputs only a partial point cloud—missing the object’s backside and fine structure. 
The fourth column visualizes Hunyuan3D-2’s result, which reconstructs a complete object mesh but fails to disentangle the hand from the object, leading to implausible fused geometry.
    }
    \label{fig:intermediate}
\end{figure}

%% file: fig/more_qualitative/supp1.tex
\begin{figure}
    \centering
    \includegraphics[width=0.8\linewidth]{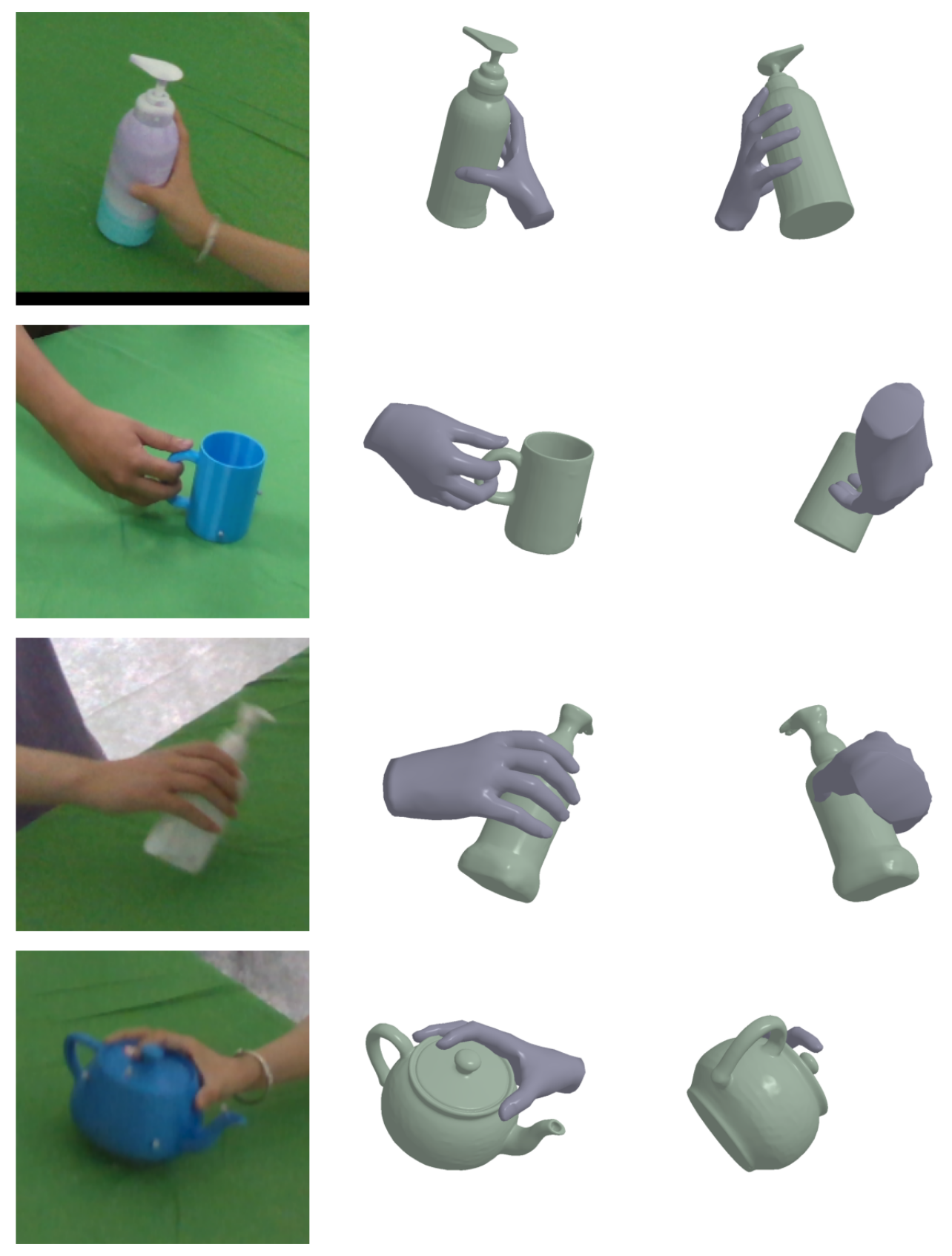}
    \caption{
    \textbf{Qualitative results on OakInk.} 
    Qualitative results of HOI reconstructions produced by our method on OakInk~\cite{yang2022oakink} dataset.
    }
    \label{fig:oakink_1}
\end{figure}

%% file: fig/more_qualitative/supp2.tex
\begin{figure}
    \centering
    \includegraphics[width=0.8\linewidth]{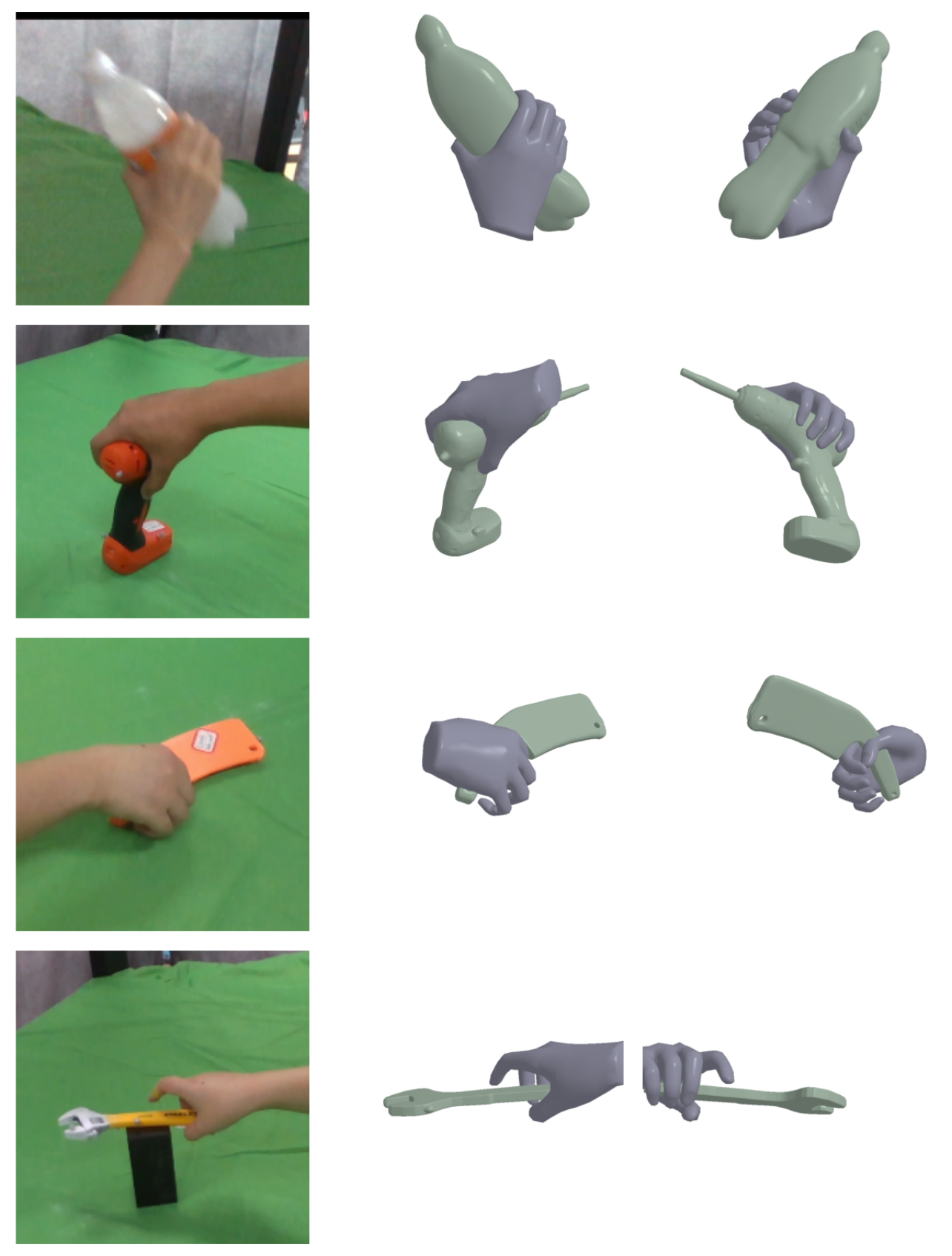}
    \caption{
    \textbf{Qualitative results on OakInk.} 
    Qualitative results of HOI reconstructions produced by our method on OakInk~\cite{yang2022oakink} dataset.
    }
    \label{fig:oakink_2}
\end{figure}

%% file: fig/more_qualitative/supp3.tex
\begin{figure}
    \centering
    \includegraphics[width=0.8\linewidth]{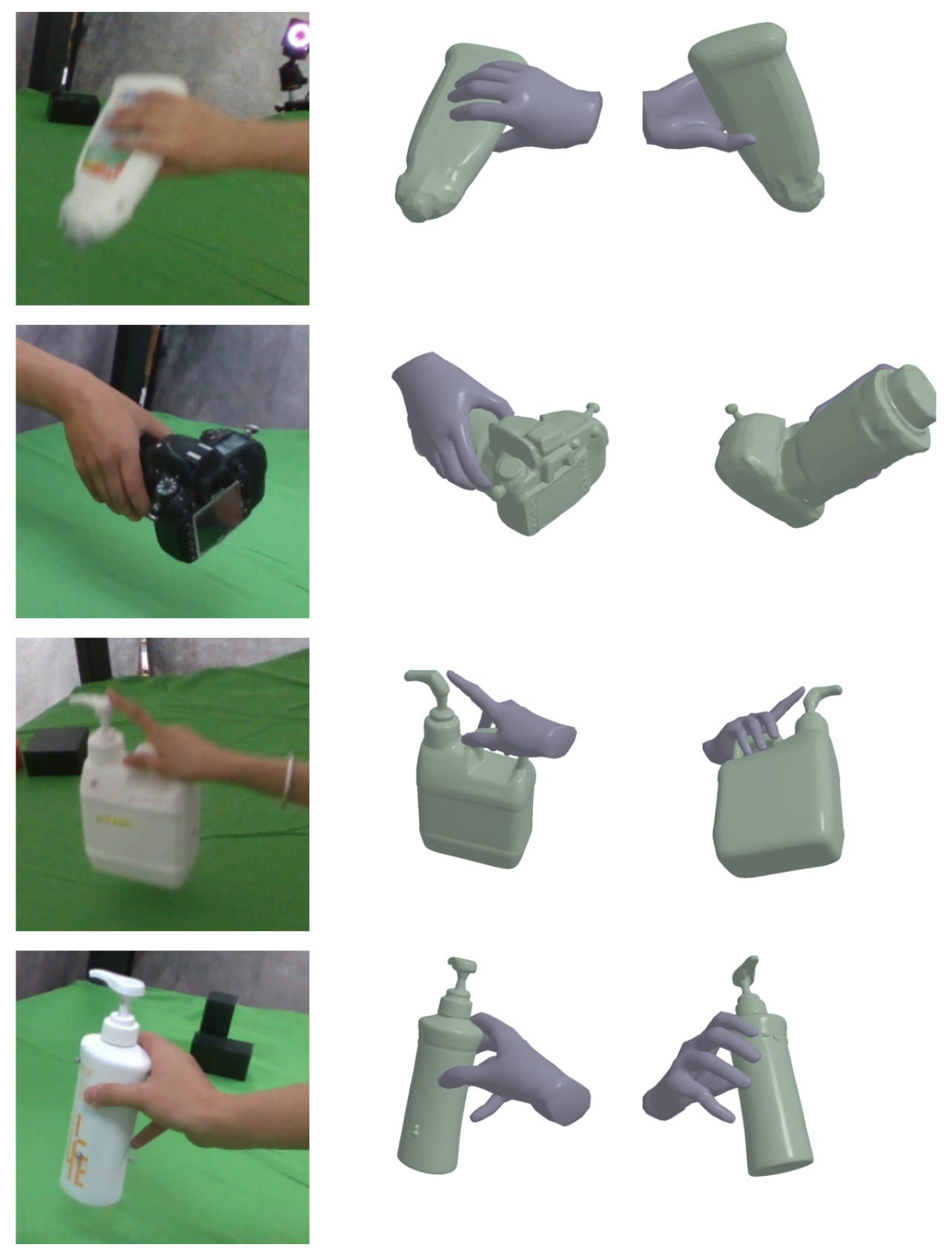}
    \caption{
    \textbf{Qualitative results on OakInk.} 
    Qualitative results of HOI reconstructions produced by our method on OakInk~\cite{yang2022oakink} dataset.
    }
    \label{fig:oakink_3}
\end{figure}

%% file: fig/more_qualitative/supp4.tex
\begin{figure}
    \centering
    \includegraphics[width=0.8\linewidth]{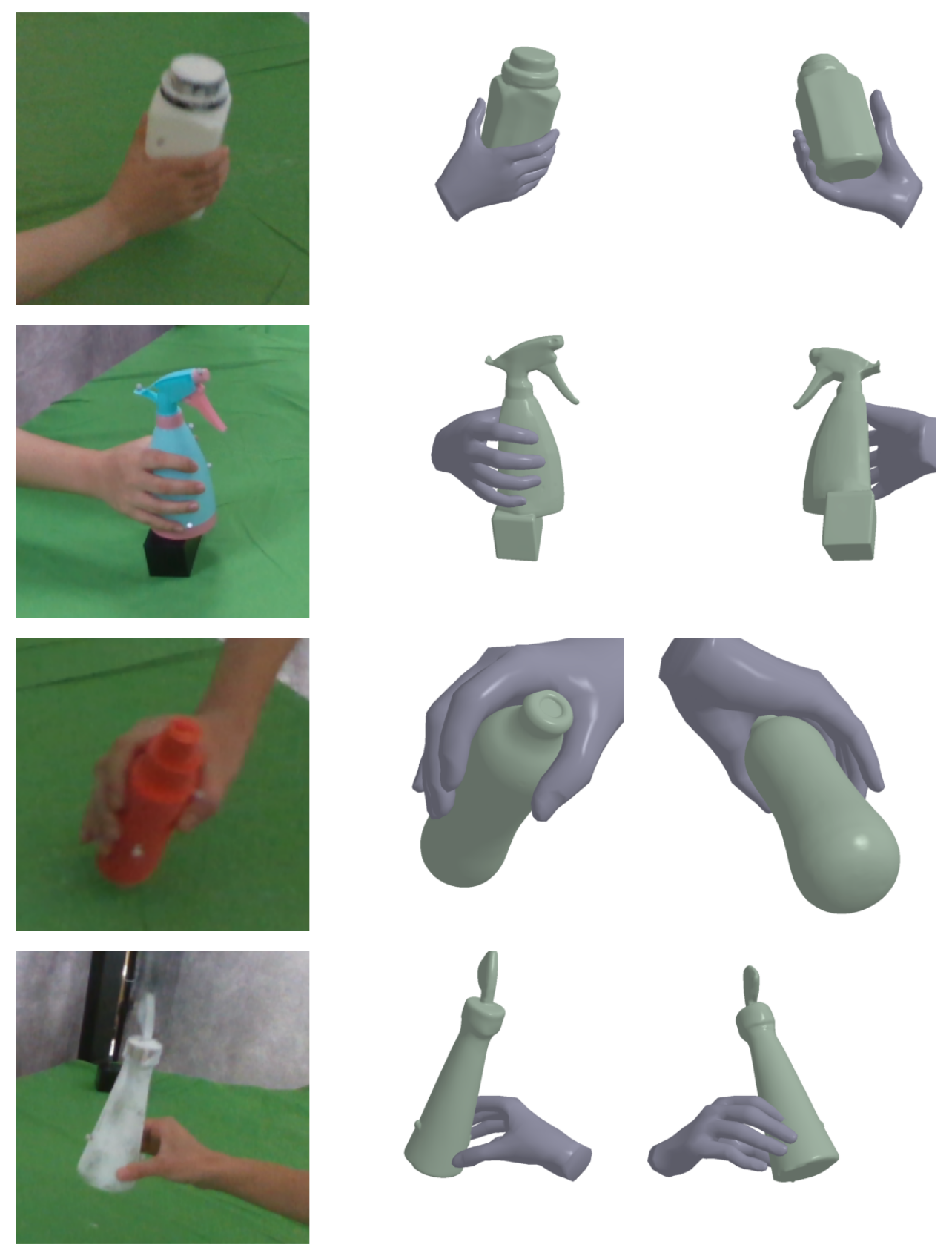}
    \caption{
    \textbf{Qualitative results on OakInk.} 
    Qualitative results of HOI reconstructions produced by our method on OakInk~\cite{yang2022oakink} dataset.
    }
    \label{fig:oakink_4}
\end{figure}

%% file: fig/more_qualitative/supp5.tex
\begin{figure}
    \centering
    \includegraphics[width=0.8\linewidth]{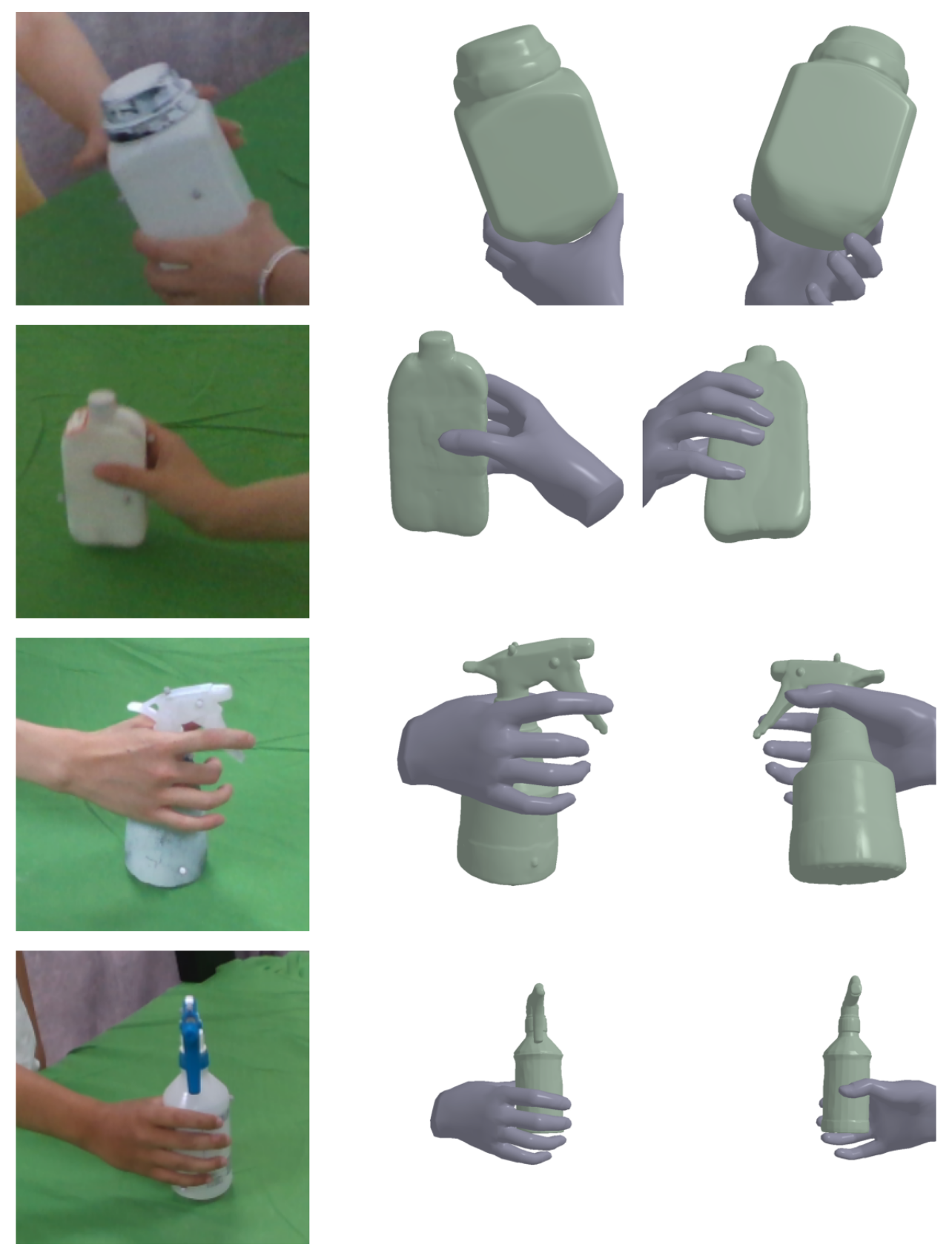}
    \caption{
    \textbf{Qualitative results on OakInk.} 
    Qualitative results of HOI reconstructions produced by our method on OakInk~\cite{yang2022oakink} dataset.
    }
    \label{fig:oakink_5}
\end{figure}

%% file: fig/more_qualitative/supp6.tex
\begin{figure}
    \centering
    \includegraphics[width=0.8\linewidth]{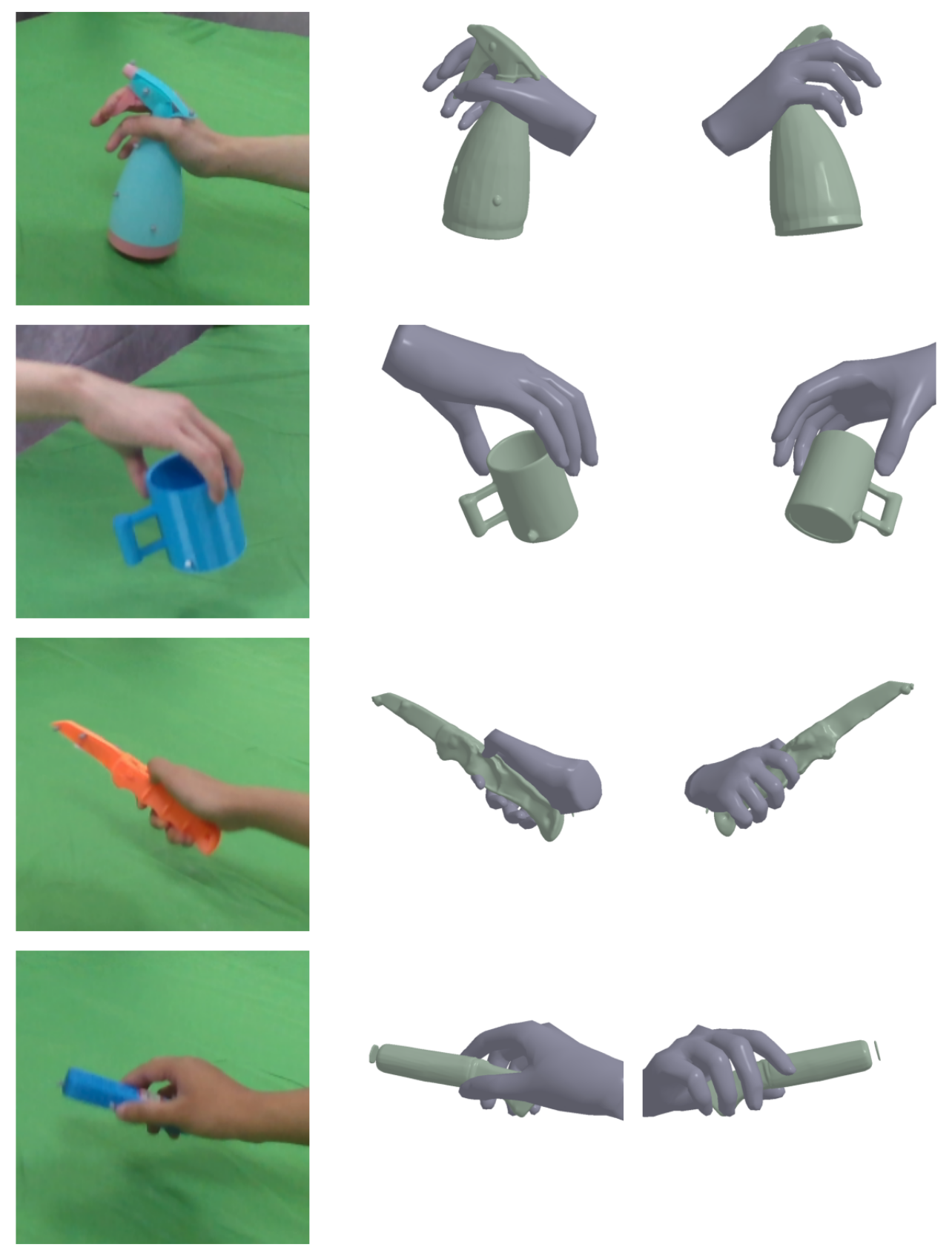}
    \caption{
    \textbf{Qualitative results on OakInk.} 
    Qualitative results of HOI reconstructions produced by our method on OakInk~\cite{yang2022oakink} dataset.
    }
    \label{fig:oakink_6}
\end{figure}

%% file: fig/more_qualitative/supp7.tex
\begin{figure}
    \centering
    \includegraphics[width=0.8\linewidth]{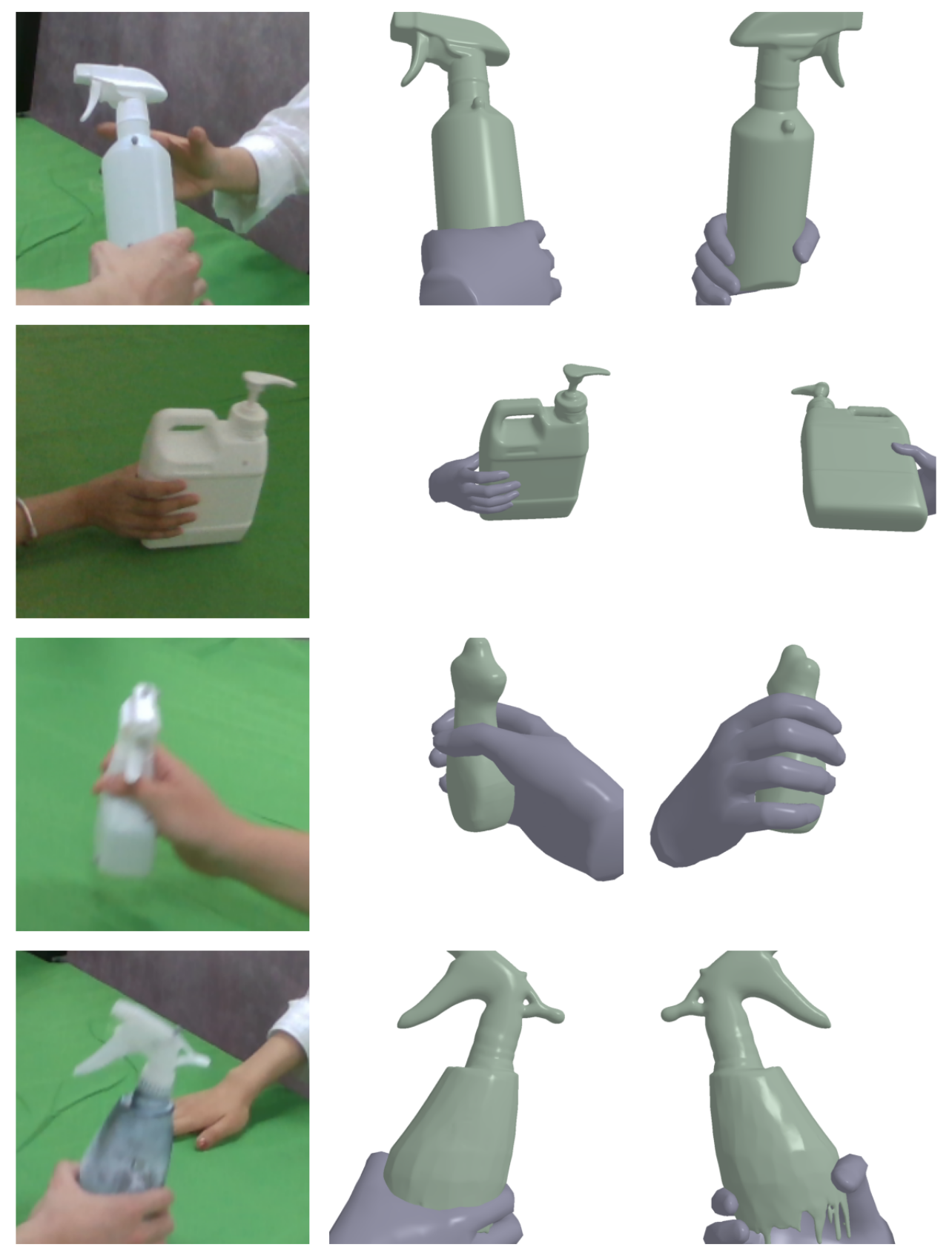}
    \caption{
    \textbf{Qualitative results on OakInk.} 
    Qualitative results of HOI reconstructions produced by our method on OakInk~\cite{yang2022oakink} dataset.
    }
    \label{fig:oakink_7}
\end{figure}

%% file: fig/more_qualitative/supp8.tex
\begin{figure}
    \centering
    \includegraphics[width=0.8\linewidth]{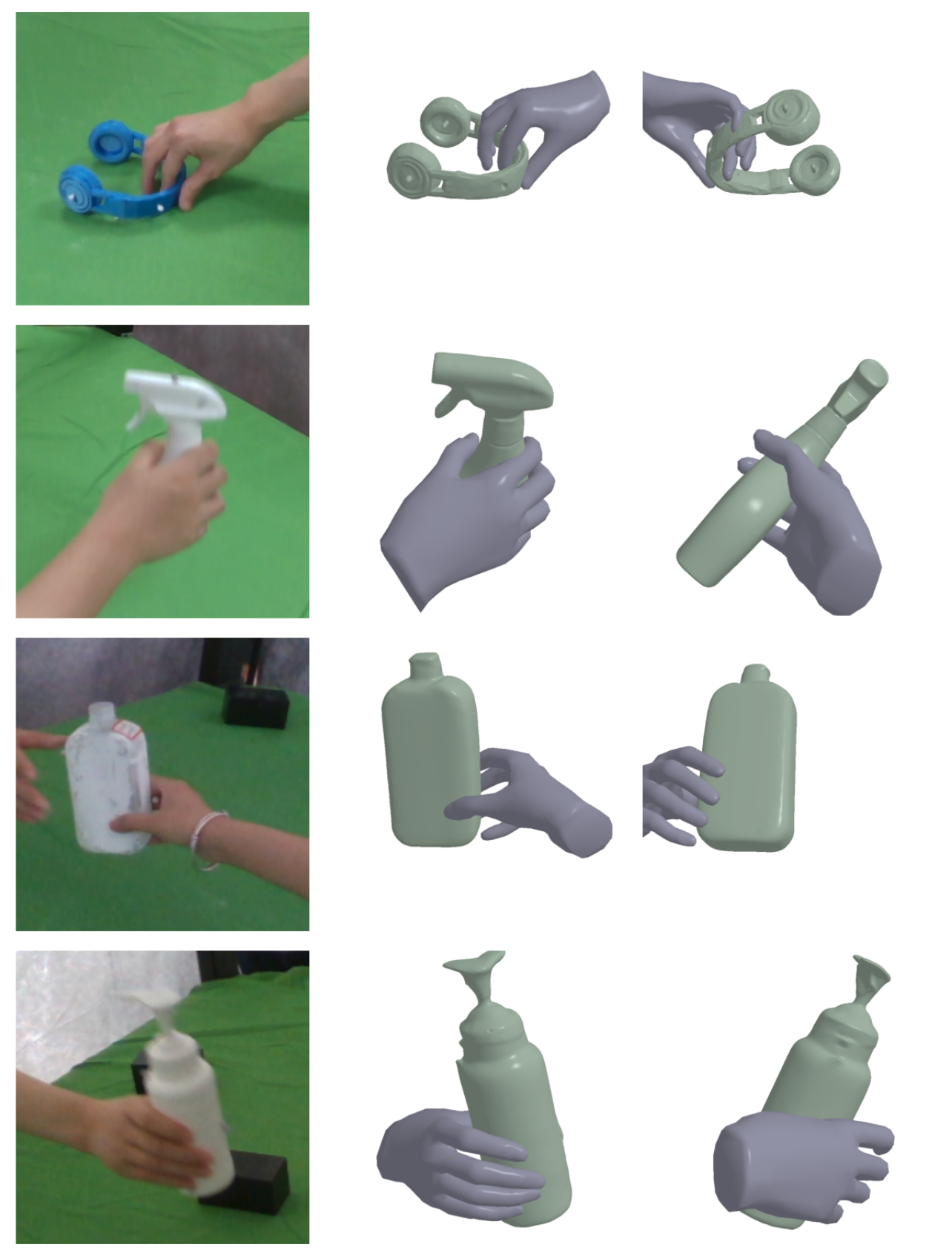}
    \caption{
    \textbf{Qualitative results on OakInk.} 
    Qualitative results of HOI reconstructions produced by our method on OakInk~\cite{yang2022oakink} dataset.
    }
    \label{fig:oakink_8}
\end{figure}

%% file: fig/more_qualitative/supp9.tex
\begin{figure}
    \centering
    \includegraphics[width=0.8\linewidth]{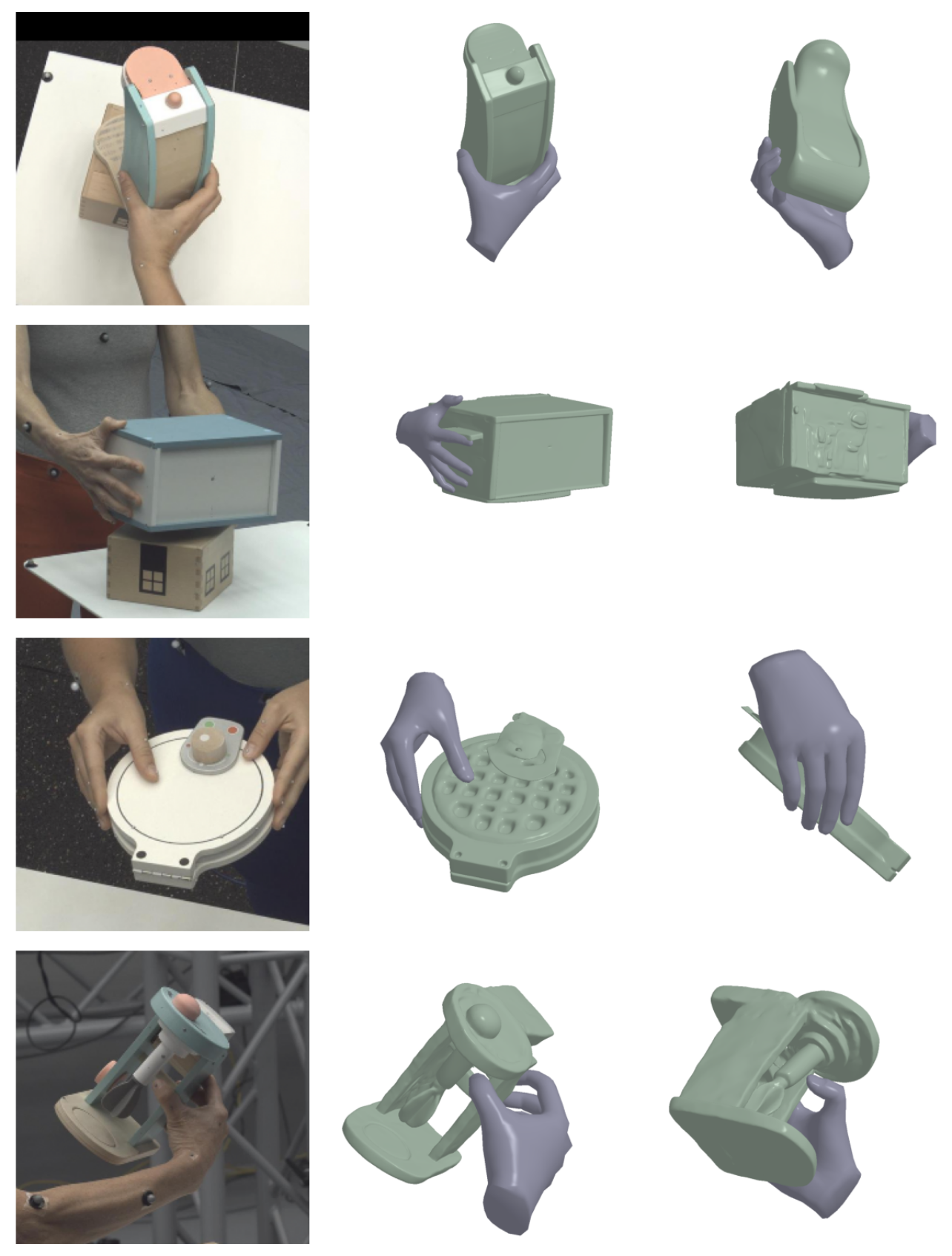}
    \caption{
    \textbf{Qualitative results on Arctic.} 
    Qualitative results of HOI reconstructions produced by our method on Arctic~\cite{fan2023arctic} dataset.
    }
    \label{fig:arctic_1}
\end{figure}

%% file: fig/more_qualitative/supp10.tex
\begin{figure}
    \centering
    \includegraphics[width=0.8\linewidth]{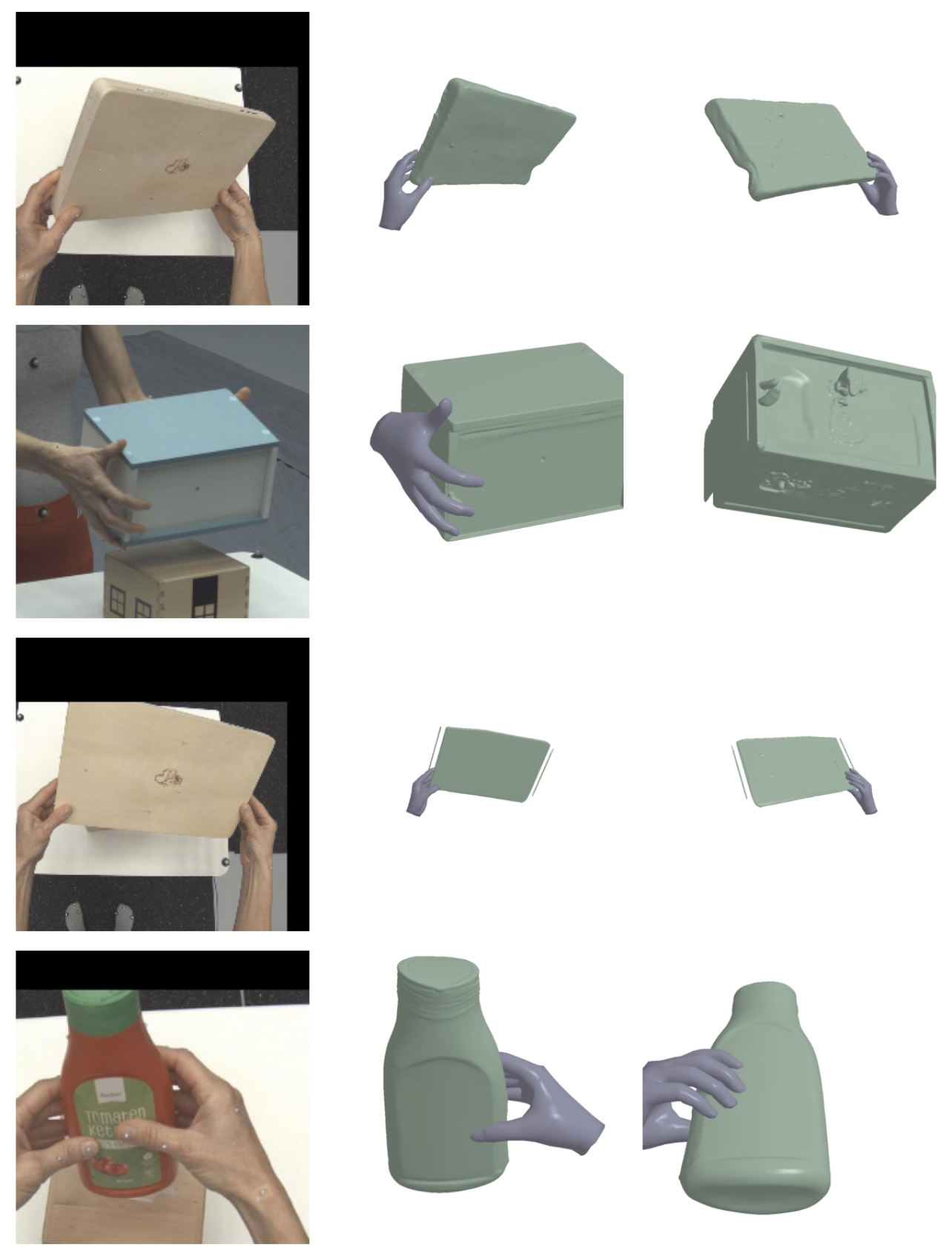}
    \caption{
    \textbf{Qualitative results on Arctic.} 
    Qualitative results of HOI reconstructions produced by our method on Arctic~\cite{fan2023arctic} dataset.
    }
    \label{fig:arctic_2}
\end{figure}

%% file: fig/more_qualitative/supp11.tex
\begin{figure}
    \centering
    \includegraphics[width=0.8\linewidth]{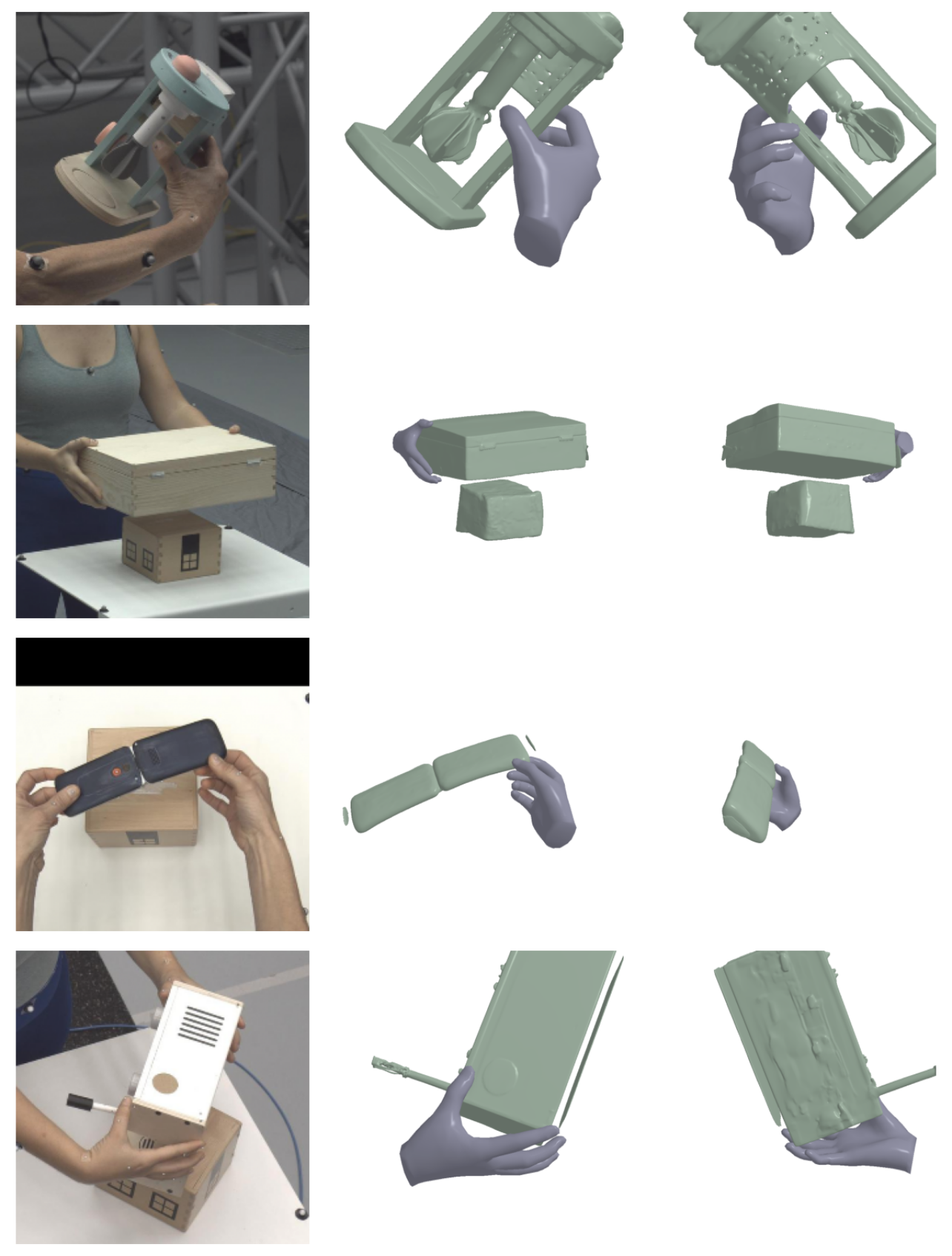}
    \caption{
    \textbf{Qualitative results on Arctic.} 
    Qualitative results of HOI reconstructions produced by our method on Arctic~\cite{fan2023arctic} dataset.
    }
    \label{fig:arctic_3}
\end{figure}

%% file: fig/more_qualitative/supp12.tex
\begin{figure}
    \centering
    \includegraphics[width=0.8\linewidth]{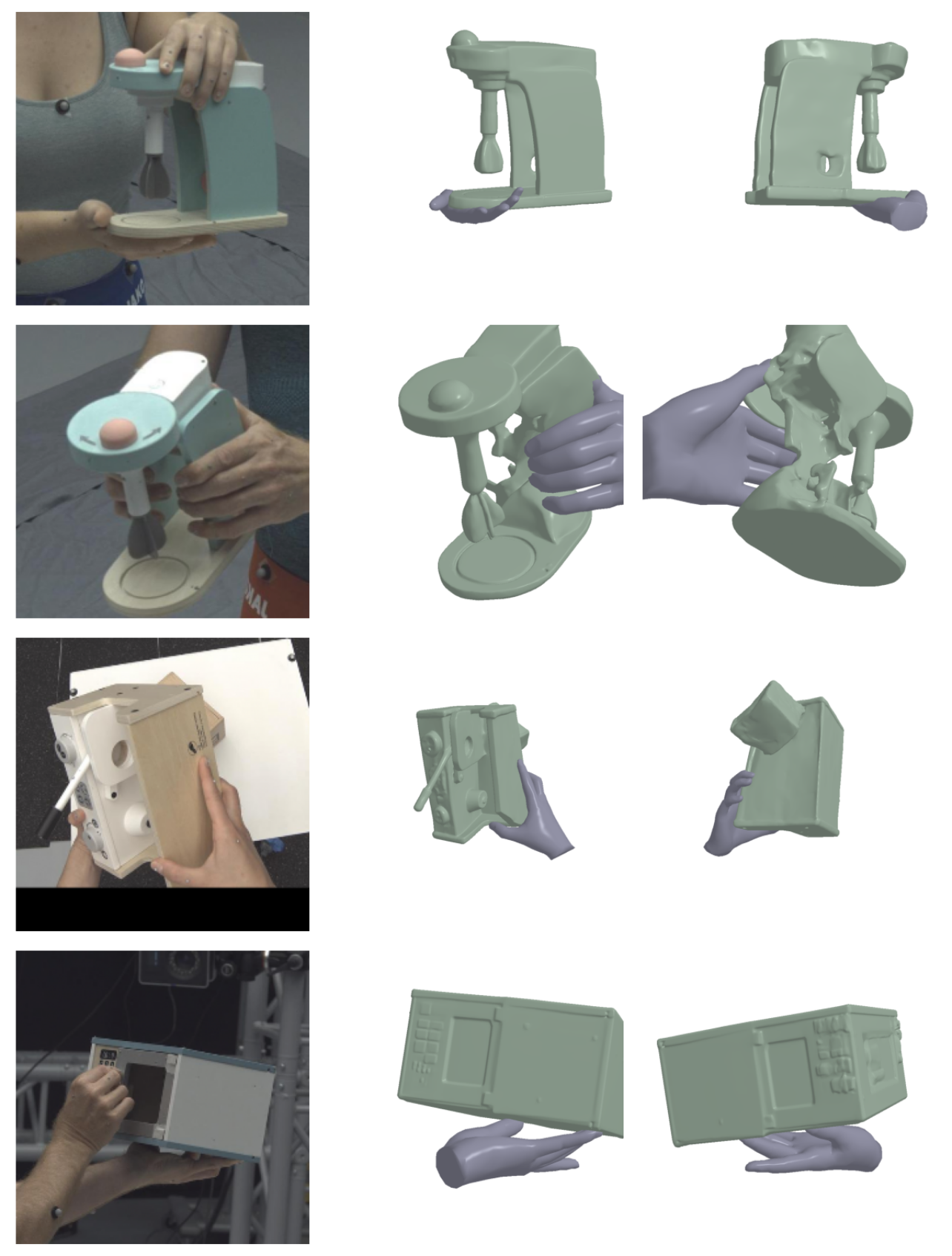}
    \caption{
    \textbf{Qualitative results on Arctic.} 
    Qualitative results of HOI reconstructions produced by our method on Arctic~\cite{fan2023arctic} dataset.
    }
    \label{fig:arctic_4}
\end{figure}

%% file: fig/more_qualitative/supp13.tex
\begin{figure}
    \centering
    \includegraphics[width=0.8\linewidth]{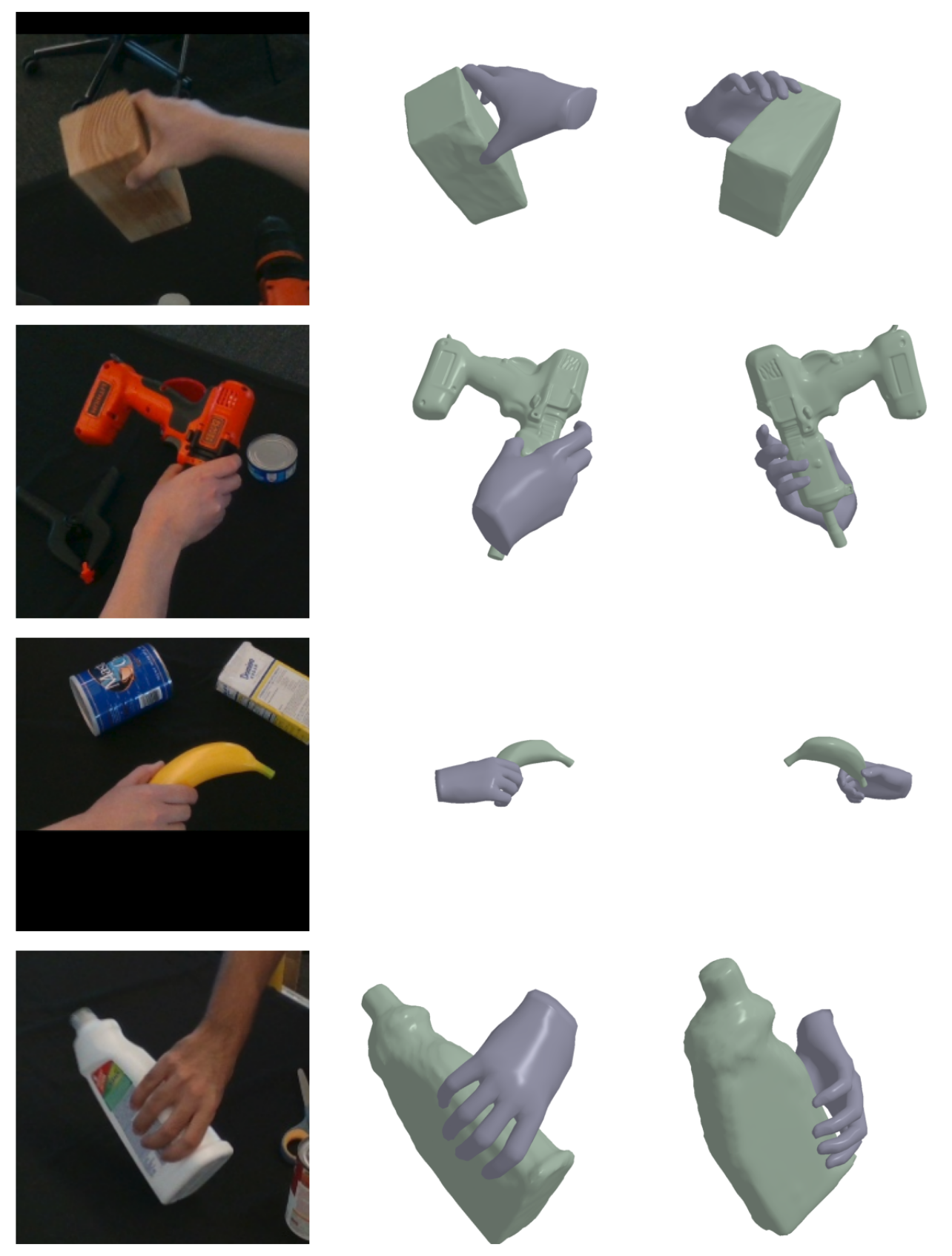}
    \caption{
    \textbf{Qualitative results on DexYCB.} 
    Qualitative results of HOI reconstructions produced by our method on DexYCB~\cite{chao2021dexycb} dataset.
    }
    \label{fig:dexycb_1}
\end{figure}

%% file: fig/more_qualitative/supp14.tex
\begin{figure}
    \centering
    \includegraphics[width=0.8\linewidth]{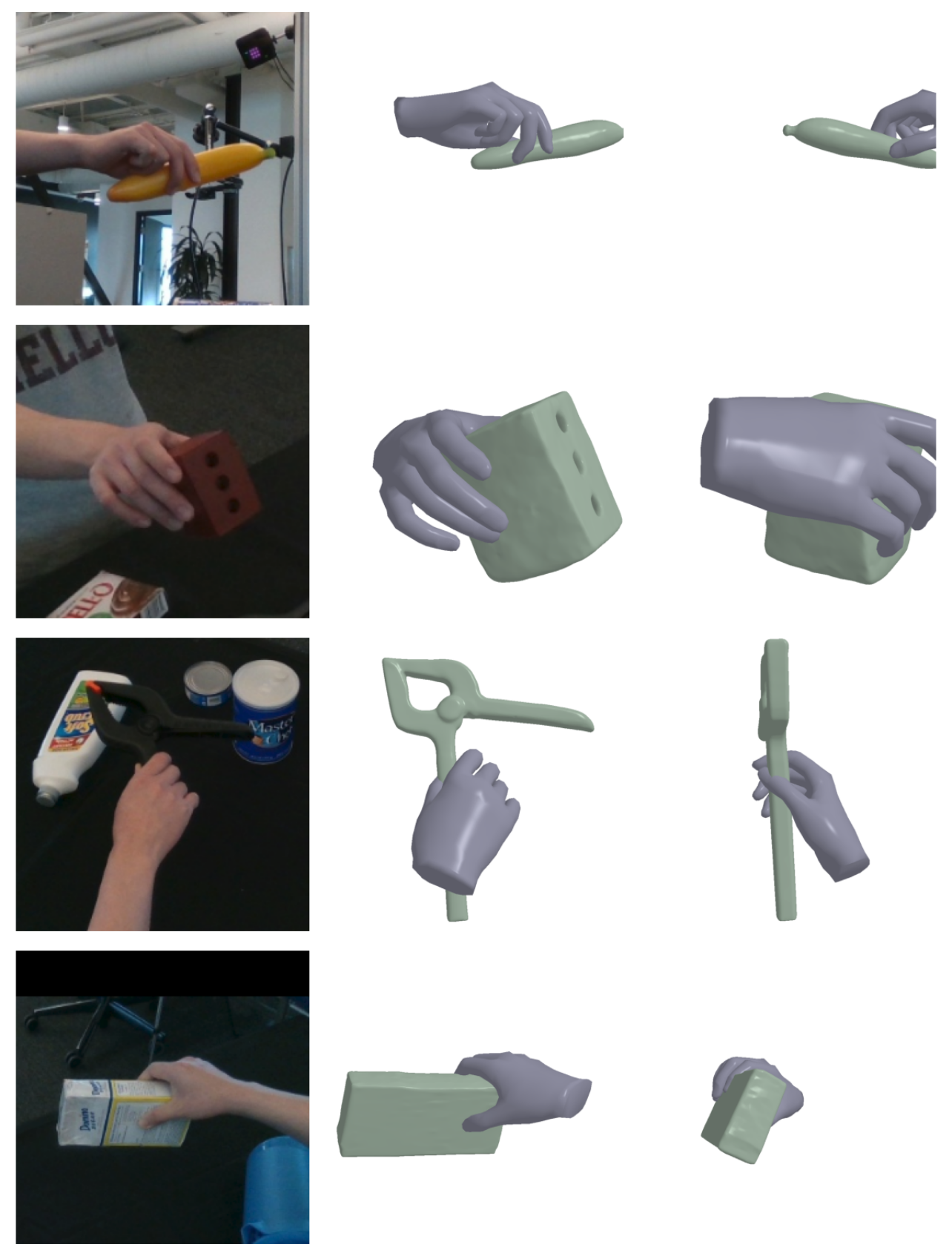}
    \caption{
    \textbf{Qualitative results on DexYCB.} 
    Qualitative results of HOI reconstructions produced by our method on DexYCB~\cite{chao2021dexycb} dataset.
    }
    \label{fig:dexycb_2}
\end{figure}